\documentclass{article}

\PassOptionsToPackage{dvipsnames,table}{xcolor}

\usepackage[accepted]{icml2026}
\usepackage{microtype}
\usepackage[most]{tcolorbox}
\usepackage{listings}
\usepackage{xcolor}
\usepackage{booktabs} % for professional tables
\usepackage{tikz}
\usepackage{subfig}
\usetikzlibrary{arrows.meta, positioning, calc, shapes.geometric, fit, backgrounds}

\usepackage{tabularx} 
\newcolumntype{Y}{>{\centering\arraybackslash}X} 
\usepackage{amssymb}
\usepackage{mathtools}
\usepackage{amsthm}
\usepackage{soul}
\usepackage{longtable}

\usepackage{algorithmic}
\usepackage{titlesec}

\usepackage[textsize=tiny]{todonotes}
\usepackage{caption}
\usepackage{multirow}
\usepackage{multicol}
\usepackage{adjustbox}

\usepackage{xspace}
\usepackage{pifont}
\usepackage{latexsym}
\usepackage{inconsolata}
\usepackage{arydshln}
\usepackage{enumitem}
\usepackage{wrapfig}
\usepackage{array}

\titlespacing\section{0pt}{8pt}{0pt}
\titlespacing\subsection{0pt}{3pt}{0pt}
\titlespacing\paragraph{0pt}{0pt}{0pt}

\theoremstyle{plain}

\theoremstyle{definition}

\theoremstyle{remark}

\newcommand{\M}{{Fox}}

\newcommand{\comt}[1]{#1}

\renewcommand{\comt}[1]{}

\definecolor{mydarkblue}{rgb}{0,0.08,0.45}
\definecolor{darkgreen}{rgb}{0.0, 0.5, 0.0} 
\definecolor{myblue}{RGB}{235,235,250}
\definecolor{lightpink}{RGB}{222, 235, 225} %lightgreen tabcolor3
\definecolor{lightblue}{RGB}{230, 235, 245} %lightblue tabcolor5
\definecolor{lightgray}{RGB}{240, 240, 240} %lightblue tabcolor5
\definecolor{darkgray}{RGB}{220, 220, 220} %lightblue tabcolor5

\definecolor{superlightred}{rgb}{0.99, 0.92, 0.92}
\definecolor{darkgreen}{RGB}{50,100,0}
\definecolor{darkred}{RGB}{200, 0, 0}

\usepackage{hyperref}
\hypersetup{
    colorlinks=true,
    linkcolor=teal,
    citecolor=mydarkblue,
    filecolor=mydarkblue,
    urlcolor=magenta
}
\usepackage[capitalize,noabbrev]{cleveref}

\theoremstyle{plain}

\icmltitlerunning{
Dismantling Pathological Shortcuts: A Causal Framework for Faithful LVLM Decoding}
\begin{document}

\twocolumn[
\icmltitle{Dismantling Pathological Shortcuts: A Causal Framework for Faithful LVLM Decoding}
% Causal Attention Mediation: Deconstructing and Mitigating Hallucinations in Large Vision-Language Models
% Severing the Confounding Path: Causal Intervention on Noisy Mediators for Hallucination Mitigation
  % It is OKAY to include author information, even for blind submissions: the
  % style file will automatically remove it for you unless you've provided
  % the [accepted] option to the icml2026 package.

  % List of affiliations: The first argument should be a (short) identifier you
  % will use later to specify author affiliations Academic affiliations
  % should list Department, University, City, Region, Country Industry
  % affiliations should list Company, City, Region, Country

  % You can specify symbols, otherwise they are numbered in order. Ideally, you
  % should not use this facility. Affiliations will be numbered in order of
  % appearance and this is the preferred way.
  \icmlsetsymbol{equal}{*}

  \begin{icmlauthorlist}
    \icmlauthor{Liu Yu}{uestc,auckland}
    \icmlauthor{Can Chen}{uestc}
    \icmlauthor{Ping Kuang}{uestc}
    \icmlauthor{Zhikun Feng}{uestc}
    \icmlauthor{Fan Zhou}{uestc}
    \icmlauthor{Gillian Dobbie}{auckland}
  \end{icmlauthorlist}

  \icmlaffiliation{uestc}{School of Information and Software Engineering, University of Electronic Science and Technology of China, Chengdu, Sichuan, China}
  \icmlaffiliation{auckland}{School of Computer Science, University of Auckland, Auckland, New Zealand}

  \icmlcorrespondingauthor{Ping Kuang}{kuangping@uestc.edu.cn}

  % You may provide any keywords that you find helpful for describing your
  % paper; these are used to populate the "keywords" metadata in the PDF but
  % will not be shown in the document
  \icmlkeywords{Large Vision-Language Models, Hallucination Mitigation, Causal Intervention, Faithful Decoding}

  \vskip 0.3in
]

% this must go after the closing bracket ] following \twocolumn[ ...

% This command actually creates the footnote in the first column listing the
% affiliations and the copyright notice. The command takes one argument, which
% is text to display at the start of the footnote. The \icmlEqualContribution
% command is standard text for equal contribution. Remove it (just {}) if you
% do not need this facility.

% Use ONE of the following lines. DO NOT remove the command.
% If you have no special notice, KEEP empty braces:
\printAffiliationsAndNotice{}  % no special notice (required even if empty)
% Or, if applicable, use the standard equal contribution text:
% \printAffiliationsAndNotice{\icmlEqualContribution}

\begin{abstract}
Large Vision-Language Models (LVLMs) exhibit sophisticated reasoning but remain susceptible to object hallucination. Deviating from the prevailing \textit{attention intensity assumption}, we reveal a deeper dynamic structural misalignment: hallucination is triggered at decision-critical steps where specific attention heads, acting as risky mediators, decouple from visual evidence to lock onto language priors. This establishes a pathological shortcut that bypasses visual grounding. 
To dismantle this, we propose \textbf{\M} (\underline{F}aithfulness and \underline{O}bservational-flow via e\underline{X}pression-rectification), a training-free inference-time framework. \M~diagnoses structural misalignment using a visual attention entropy probe to localize risky mediators unsupervisedly. We then execute a targeted causal intervention via numerical logit saturation to physically sever the shortcut path. Finally, a conflict-gated cooperative decoding strategy reconciles interventional faithfulness with observational fluency. 
Extensive experiments demonstrate that \M~achieves SOTA performance, outperforming SID by $29.1\%$ while preserving linguistic richness. Code is available at \url{https://github.com/Cc2021start/Fox}.
\end{abstract}
\vspace{-4mm}

\begin{figure}[t]
    \centering
    \includegraphics[width=1\linewidth]{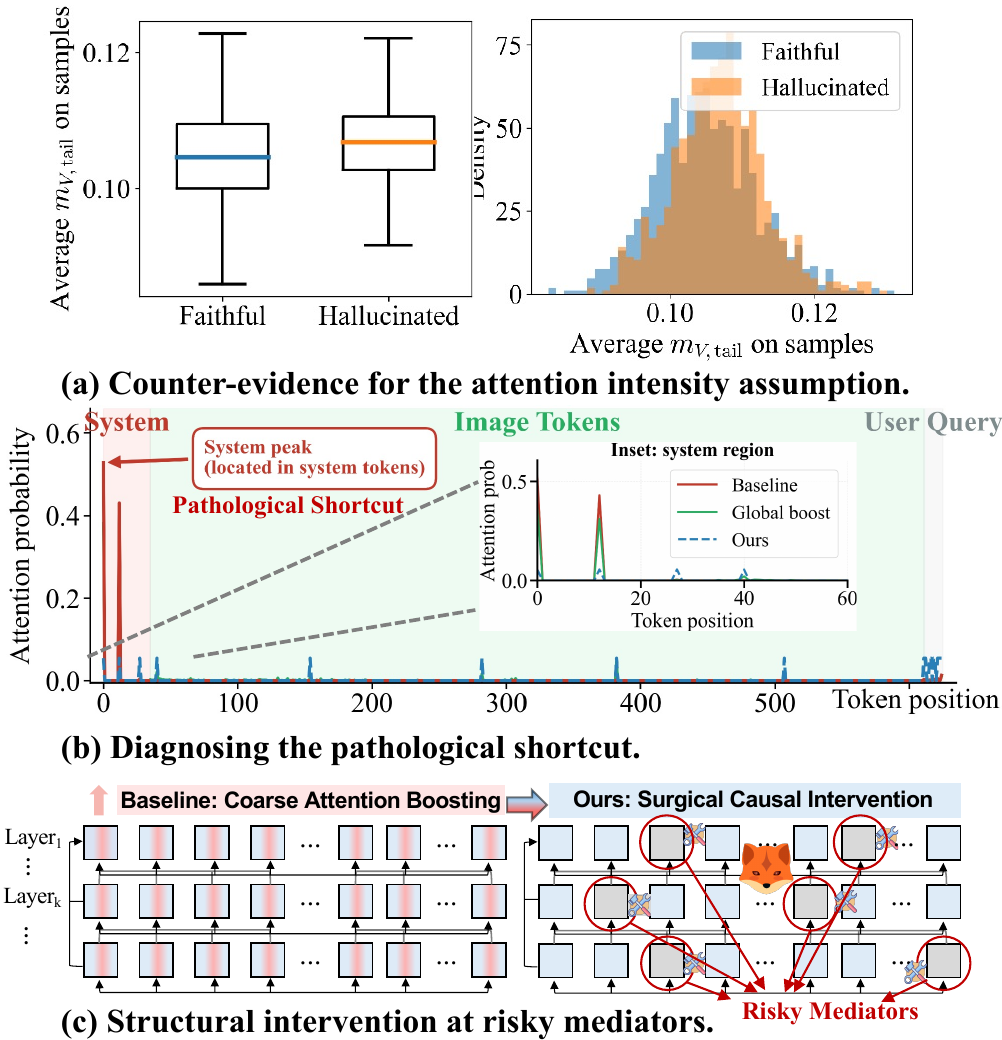}
    \caption{
    \textbf{Motivation of our work.} 
    \textbf{(a)} Global visual attention magnitude $m_{V,tail}$ and distribution lack discriminative power to identify hallucination. 
    \textbf{(b)} While global magnitude boosting (Green) fails to suppress the pathological peak on system instructions at decision-critical steps, our structural intervention (Blue) on \textit{risky mediators} eliminates this shortcut, restoring visual grounding.
    \textbf{(c)} Unlike coarse-grained enhancement across all layers (Left), \M~performs a sparse, surgical intervention on diagnosed risky mediators (Right), physically severing the prior-driven shortcut.
    }
    \label{fig:intro}
    \vspace{-4mm}
\end{figure}

\section{Introduction}
\label{sec:intro}

Large Vision-Language Models (LVLMs) have demonstrated remarkable capabilities in multimodal reasoning~\cite{liu2023visual,wan2025instructparttaskorientedsegmentationinstruction}. Despite these advancements, they frequently suffer from object hallucination---generating content that contradicts visual evidence~\cite{leng2024cursemultimodalitiesevaluatinghallucinations,nie2025mmrelbenchmarkingrelationunderstanding}. This poses severe risks in safety-critical domains, such as medical imaging or embodied AI~\cite{wang2023chatcadinteractivecomputeraideddiagnosis,tian2024drivevlmconvergenceautonomousdriving}, where a single hallucinated token can trigger catastrophic reasoning failures.

Current mitigation strategies generally fall into training-time alignment~\cite{bai2025hallucinationmultimodallargelanguage,liu2024mitigatinghallucinationlargemultimodal} or inference-time intervention~\cite{VCD,an2025mitigating,li2025mitigatinghallucinationlargevision,fazli2025mitigatinghallucinationlargevisionlanguage}. While training-based methods incur substantial computational costs, inference-time interventions have gained traction for their model-agnostic efficiency~\cite{zhang2025selfcorrectingdecodinggenerativefeedback,chen2024halcobjecthallucinationreduction,che2025hallucinatoryimagetokenstrainingfree}. Despite technical variations, most existing approaches share a common premise, which we term the \textit{attention intensity assumption}: hallucination is primarily attributed to a quantitative deficit in visual attention~\cite{chen2025mixturedecodingattentioninspiredadaptive}. Consequently, these methods seek to rectify failures by mechanically amplifying visual signals (e.g., PAI~\cite{liu2024paying}) or suppressing language priors~\cite{VCD}.
However, this intuition proves empirically incomplete, particularly for strategies predicated on global magnitude enhancement. As shown in Fig.~\ref{fig:intro}(a), $m_{V,tail}$ denotes the visual attention magnitude, representing the total attention weight allocated to image tokens. A controlled analysis reveals no statistically significant reduction in global visual attention magnitude for hallucinated outputs, with their distributions largely overlapping. This lack of discriminative power suggests that focusing solely on intensity overlooks the underlying structural misalignment of hallucination. The decisive failure is therefore not only how much visual mass is assigned, but where the final prediction is routed at the moment of content generation. More details \textit{cf.} Appendix~\ref{app:stat_setup} and \ref{app:global_diff}.

Motivated by this, we shift our focus from global magnitude to the transient pathology triggered at decision-critical steps. We observe that hallucination is driven by specific attention heads, i.e., \textit{risky mediators}, that functionally decouple from visual evidence precisely when the model commits to content-bearing generation. As depicted in Fig.~\ref{fig:intro}(b), a naive global boosting strategy (e.g., PAI~\cite{liu2024paying}) succeeds in increasing the total attention volume but fails to dismantle the localized, pathological peak on system instructions. This persistent structural bias establishes a shortcut where latent language priors bypass visual grounding to dominate the output. From a causal perspective, these heads act as unreliable mediators that reroute influence via spurious dependencies. As shown in Fig.~\ref{fig:intro}(c), addressing this requires a shift from uniform, token-level adjustments toward sparse, head-level causal interventions.

To dismantle this pathological structure, we propose \M~(\underline{F}aithfulness and \underline{O}bservational-flow via e\underline{X}pression-rectification), a training-free framework grounded in a Structural Causal Model (SCM). We reformulate decoding as a causal process where attention heads at specific decision-critical steps serve as mediators. Specifically, we introduce visual attention entropy as an unsupervised probe to pinpoint risky mediators exhibiting high visual uncertainty. Upon detection, we execute a targeted intervention via the $\mathbf{do}$-operator—implemented as numerical logit saturation—to physically sever the shortcut path, forcing the model to rely on direct visual evidence. Finally, to reconcile interventional faithfulness with linguistic fluency, we implement a conflict-gated cooperative decoding strategy that dynamically fuses observational and interventional distributions. Our main contributions are summarized as follows:
\begin{itemize}[itemsep=0pt,topsep=0pt,parsep=0pt,leftmargin=*]
    \item We challenge the prevailing attention intensity assumption by revealing that hallucination stems from dynamic structural misalignment. We identify risky mediators---sparse heads that structurally disconnect from visual inputs at decision-critical steps---offering a novel mechanistic perspective on LVLM failures.
    
    \item We propose \M, a principled inference-time framework rooted in SCM. By intersecting decision-critical steps with visual attention entropy probes, we achieve precise, unsupervised localization and $\mathbf{do}$-driven suppression of pathological shortcuts.
    
    \item Extensive experiments demonstrate that \M~significantly outperforms existing baselines, achieving a 22.9\% improvement on CHAIR and mitigating hallucination while preserving descriptive richness.
\end{itemize}

\begin{figure*}[ht]
    \centering
    \includegraphics[width=0.85\linewidth]{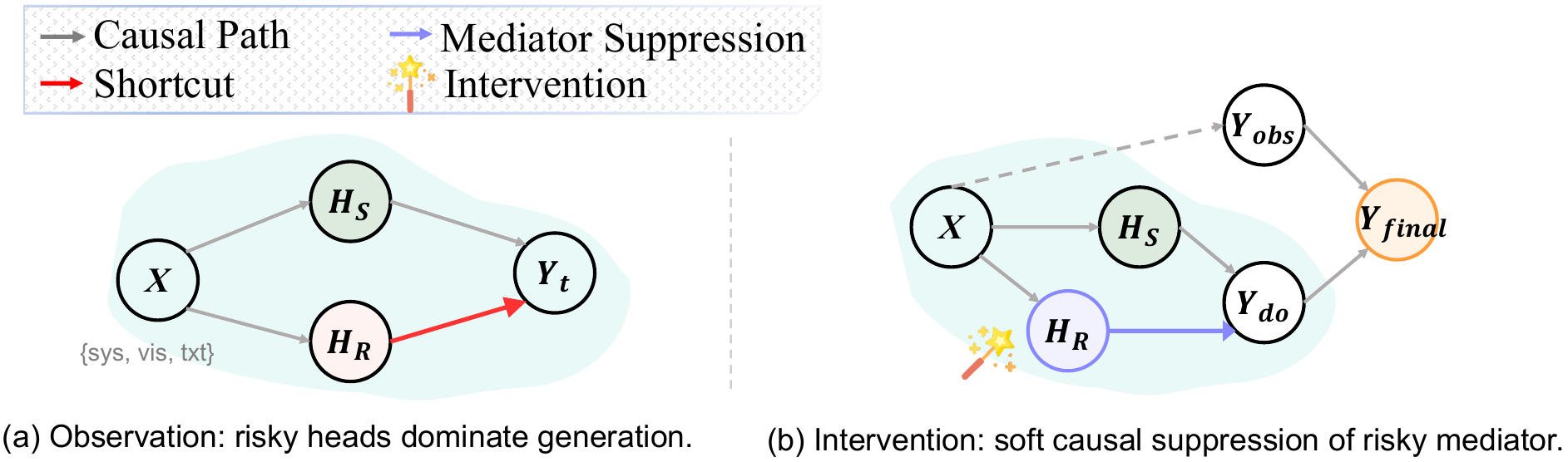} 
    \caption{\textbf{Structural Causal Model (SCM) of the LVLM Decoding Path.} 
    (a) \textbf{Observational SCM:} The latent mediators $H$ are localized at decision-critical steps. While stable mediators $H_S$ maintain visual grounding, risky mediators $H_R$ trigger a \textbf{pathological shortcut} (red arrow) from language priors $\mathbf{X}_{sys}$ to output $Y_t$. 
    (b) \textbf{Interventional SCM:} By applying $\mathbf{do}(H_R)$, we sever the shortcut. The final output is dynamically reconciled from observational ($P_{obs}$) and interventional ($P_{do}$) distributions.}
    \label{fig:scm}
\end{figure*}

\vspace{-3mm}
\section{Related Work}

\noindent\textbf{Hallucination Mitigation in LVLMs.}
Existing strategies generally fall into training-time alignment~\cite{sun2023aligninglargemultimodalmodels,zhou2024analyzingmitigatingobjecthallucination} and inference-time intervention~\cite{zhualleviating,tong2025mitigating,yu2026causally}. 
Given the substantial cost of retraining, recent research has favored inference-time methods: contrastive decoding approaches such as VCD~\cite{VCD} and ICD~\cite{ICD} penalize language priors via negative constraints, while reweighting methods~\cite{liu2024paying,zou2024look} mechanically amplify visual signals. 
More granular studies, such as OPERA~\cite{huang2024opera} and SEVI~\cite{zhao2025mitigatinghallucinationlargevisionlanguage}, attempt to regulate generation by penalizing over-trust or emphasizing specific semantic layers.
However, they predominantly hinge on the attention intensity assumption, treating the magnitude of attention as the primary proxy for faithfulness. 
Unlike these intensity-based heuristics, we argue that hallucination stems from a dynamic structural misalignment. By shifting the focus from global magnitude to the visual attention entropy at decision-critical steps, we distinguish valid reasoning from confident but misaligned hallucinations, offering a more precise diagnostic granularity.

\noindent\textbf{Causal Inference in Multimodal Reasoning.}
Structural Causal Models (SCMs) provide a rigorous framework for debiasing and interpretability in vision-language tasks~\cite{pearl2009causality}. 
Related efforts use invariant learning, mixup, generated sentences, and graph contrastive pre-training to mitigate bias or enrich pretrained models~\cite{zhou2023causal,mao2023debiasing,yu2023mixup,yu2024biases,yu2025bridging,yu2025amplifying}, while bimodal debiasing extends this principle to text-to-image generation~\cite{yu2025bimodal}.
Recent studies like CausalMM~\cite{CausalMM} and \citet{huang2024brings} employ SCMs to analyze hallucinations, typically utilizing input-level counterfactuals---such as masking image regions or tokens---to estimate causal effects.
While effective for post-hoc diagnosis, these input-level perturbations are often too coarse to rectify the model's internal reasoning dynamics. 
In contrast, we reformulate internal attention heads as dynamic mediators. 
This allows us to perform surgical interventions via the $\mathbf{do}$-operator directly on the latent information flow when the model reaches decision-critical queries. 
By physically severing pathological shortcut paths within the network rather than altering external inputs, \M~achieves a principled, training-free restoration of visual grounding.

% \vspace{-3mm}
\section{Preliminary}
\label{sec:preliminary}

\noindent\textbf{Problem Formulation.} 
We consider an LVLM $\mathcal{F}_{\theta}$ that processes a multimodal input $\mathbf{X}$ partitioned into three semantic subspaces: visual $\mathbf{X}_{vis}$ (indices $\mathcal{I}_{vis}$), system instructions $\mathbf{X}_{sys}$ ($\mathcal{I}_{sys}$), and textual history $\mathbf{X}_{txt}$ ($\mathcal{I}_{txt}$). 
The model generates $Y=\{y_1,\dots,y_L\}$ autoregressively, where the next-token probability is $P(y_t \mid \mathbf{X}, y_{<t}) = \text{Softmax}(\mathcal{F}_{\theta}(\mathbf{X}, y_{<t}))$.
The internal routing mechanism is driven by multi-head attention. For head $h$ at layer $l$, the attention logits $\mathbf{L}^{(l,h)}$ and weights $\mathbf{A}^{(l,h)}$ are computed as:
\begin{equation}
    \mathbf{L}^{(l,h)} = \frac{\mathbf{Q}^{(l,h)} (\mathbf{K}^{(l,h)})^\top}{\sqrt{d_k}}; \quad \mathbf{A}^{(l,h)} = \text{Softmax}(\mathbf{L}^{(l,h)}). \nonumber
\end{equation}
Our causal intervention (§~\ref{sec:method}) directly modulates $\mathbf{L}^{(l,h)}$ to rectify the information flow before normalization.

\noindent\textbf{Causal Formulation.} 
We formalize the decoding process as an SCM in Fig.~\ref{fig:scm}. We identify the attention heads acting at decision-critical steps $\mathcal{Q}$ as the dynamic mediators $H$, which transmit causal influence from inputs to output. 
We posit that causal mediation is temporally sparse, where information is aggregated at specific nodes $(q \in \mathcal{Q}, h \in \mathcal{H})$ rather than uniformly across tokens. 
Ideally, faithful generation requires $H$ to reliably transmit evidence via the visual path: $\mathbf{X}_{vis} \to H \to Y_t$.

However, we observe a structural misalignment of mediation in Fig.~\ref{fig:scm}(a), where mediators bifurcate into: 
(1) Stable mediators $H_S$, maintaining grounded visual attention; 
(2) Risky mediators $H_R$, which functionally decouple from visual evidence to lock onto language priors. 
Specifically, $H_R$ establishes a pathological shortcut: $\mathbf{X}_{sys} \to H_R \to Y_t$. 
While any text carries priors, $\mathbf{X}_{sys}$ (e.g., ``You are a helpful assistant'') serves as the primary anchor for latent priors when visual grounding fails, leading to object hallucination.

\noindent\textbf{Causal Intervention.} 
To block this shortcut without retraining, we apply the intervention $\mathbf{do}(H_R := \text{noise})$. 
This operation suppresses $H_R$, forcing the model to rely on the faithful visual path. 
This process yields two distributions: the observational anchor $P_{obs}$ and the interventional candidate $P_{do}$. 
As shown in Fig.~\ref{fig:scm}(b), the final output $Y_{\text{final}}$ is derived via a conflict-gated causal fusion that dynamically reconciles these branches:
\begin{equation}
    Y_{\text{final}} \sim \text{Softmax}(f_{\text{gate}}(\mathbf{z}_{obs}, \mathbf{z}_{do})).
\end{equation}
The overall framework is guided by three research questions (RQs) targeting diagnosis, intervention, and cooperation.
The Algorithm and detailed methodology \textit{cf.} Appendix~\ref{sec_app:method}.
\section{Method}
\label{sec:method}

\subsection{Causal Diagnosis of Risky Mediators}
\label{sec:diagnosis}

\textbf{RQ1.}
\textit{Given the dense multi-head attention mechanism in LVLMs, how can we unsupervisedly identify the specific attention heads that facilitate the pathological shortcut $\mathbf{X}_{sys} \to H_R \to Y_t$?}

\begin{tcolorbox}[
  tile, size=fbox, boxsep=2mm, boxrule=0pt, top=0pt, bottom=0pt,
  borderline west={1mm}{0pt}{purple!70!pink!30}, 
  colback=yellow!10!white, 
]
\textit{\textbf{\textcolor{gray!50!black}{Insight I:}}}
Hallucination is a dynamic structural misalignment rather than a uniform signal deficit. From a causal perspective, the breakdown of visual grounding is concentrated at specific nodes where the model aggregates multimodal context to update its internal states. Diagnosing hallucination thus requires dual-axis localization: identifying the intersection of decision-critical steps (temporal axis) and risky mediators (spatial axis).
\end{tcolorbox}

\noindent\textbf{Temporal Axis: Identifying Decision-Critical Steps.}
In autoregressive transformers, information flow is anisotropic and temporally sparse. We posit that the shortcut mechanism is most detectable at decision-critical steps $\mathcal{Q}$, where the model transitions from prompt encoding to token synthesis. Unlike prior methods that average signals across all tokens, we pinpoint two pivotal temporal anchors derived from the input $\mathbf{X}$ and history $y_{<t}$:
\begin{itemize}[leftmargin=*,itemsep=0pt,topsep=0pt,parsep=0pt]
    \item \textbf{The Multimodal Handshake ($\mathbf{x}_{\text{last}} \in \mathbf{X}$):} The terminal token of the prefix sequence $\mathbf{X}$. As the final node of multimodal integration, $\mathbf{x}_{\text{last}}$ acts as the aggregator that compresses the visual-linguistic context ($\mathbf{X}_{vis}, \mathbf{X}_{sys}$) into the initial hidden state. A failure to ground on $\mathbf{X}_{vis}$ here leads to a corrupted trajectory initialization.
    \item \textbf{The Autoregressive Anchor ($y_{t-1}$):} The immediate predecessor of the current prediction $y_t$. As the proximal causal parent in the SCM, $y_{t-1}$ serves as the most sensitive probe for immediate prior dominance by language-biased heads during the generative phase.
\end{itemize}

\begin{figure}[t]
    \centering
    \includegraphics[width=1\linewidth]{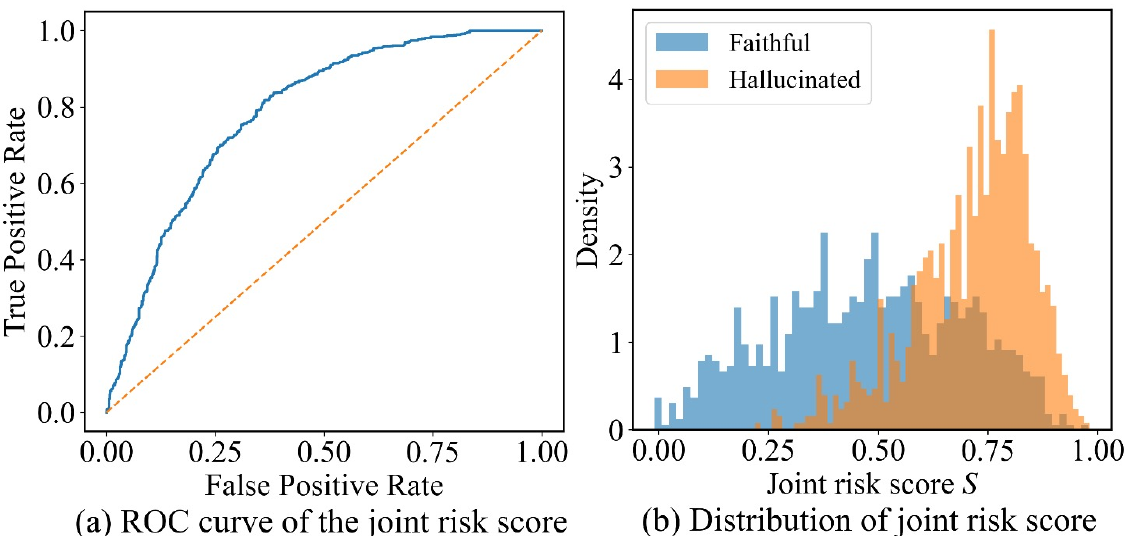}
    \caption{
        \textbf{Empirical validation of the joint risk score ($S$).} 
        \textbf{(a) Diagnostic fidelity:} The ROC curve demonstrates that the aggregated joint risk score reliably distinguishes hallucinated trajectories from faithful ones (AUC=0.818). 
        \textbf{(b) Structural decoupling:} The distribution shift confirms that hallucination (orange) is characterized by higher joint risk, signifying the concurrent collapse of visual reliability and activation of language priors. 
        \textbf{Score construction:} The sample-level score is an aggregation of head-level joint risk $S^{(l,h)}$ weighted by their respective causal contributions (Eq.~\eqref{eq:score}).
    }
    \label{fig:entropy}
    \vspace{-3mm}
\end{figure}

\noindent\textbf{Spatial Axis: Quantifying Structural Shortcuts.}
At the identified steps $\mathcal{Q}$, we inspect the latent mediators (attention heads) to detect structural decoupling. Following the subspaces defined in \S~\ref{sec:preliminary}, we partition the key space into system ($\mathcal{I}_{sys}$), visual ($\mathcal{I}_{vis}$), and text ($\mathcal{I}_{txt}$) indices. We quantify the prior-path activation for head $h$ at layer $l$ by computing the system attention magnitude at $\mathcal{Q}$:
\begin{equation}
    m_{\mathrm{sys}}^{(l,h)} = \frac{1}{|\mathcal{Q}|} \sum_{q \in \mathcal{Q}} \sum_{k \in \mathcal{I}_{sys}} \mathbf{A}^{(l,h)}_{q,k},
\end{equation}
where $m_{\mathrm{sys}}$ measures the reliance on $\mathbf{X}_{sys}$ relative to the current multimodal context.  More details of  $m_{\mathrm{sys}}$ \textit{cf.} Appendix~\ref{app:sys_hallucination}.

\noindent\textbf{Unsupervised Diagnosis via Visual Attention Entropy.}
A shortcut is only verified when the reliance on priors occurs alongside the collapse of visual reliability. To measure the causal uncertainty of the visual pathway $\mathbf{X}_{vis} \to H$, we re-normalize the attention weights strictly over the visual indices $\mathcal{I}_{vis}$ to obtain the local distribution $\hat{\mathbf{A}}^{(l,h)}_{q}$:
\begin{equation}
\begin{aligned}
    \hat{\mathbf{A}}_{q,j}^{(l,h)} &= \frac{\mathbf{A}_{q,j}^{(l,h)}}{\sum_{k \in \mathcal{I}_{vis}} \mathbf{A}_{q,k}^{(l,h)}}, \quad j \in \mathcal{I}_{vis}; \\
    H_{\mathrm{vis}}^{(l,h)} &= \frac{1}{|\mathcal{Q}|} \sum_{q \in \mathcal{Q}} \text{Entropy}(\hat{\mathbf{A}}_{q}^{(l,h)}).
\end{aligned}
\end{equation}
High $H_{\mathrm{vis}}$ signifies attentional dispersion, where the mediator fails to extract grounded evidence (further visualization in Appendix~\ref{app:visual_uncertainty}).

\noindent\textbf{Joint Risk Scoring and Mediator Selection.}
We define the risky mediator $H_R$ as a node that facilitates the pathological shortcut through two concurrent conditions: (1) high causal uncertainty $H_{\mathrm{vis}}$ and (2) active prior-path transmission $m_{\mathrm{sys}}$. The joint risk score is:
\begin{equation}
    S^{(l,h)} = m_{\mathrm{sys}}^{(l,h)} \cdot H_{\mathrm{vis}}^{(l,h)}. \label{eq:score}
\end{equation}
This multiplicative form functions as a conjunctive filter: it selectively pinpoints mediators where the breakdown of visual grounding directly correlates with the dominance of language priors. This avoids over-penalizing heads that remain visually grounded, making the score more transferable across LVLM backbones with different attention scales. As evidenced by Fig.~\ref{fig:entropy}, $S$ serves as a robust diagnostic probe to localize $H_R$. 

\begin{figure*}[htbp]
\centering
\includegraphics[width=1\linewidth]{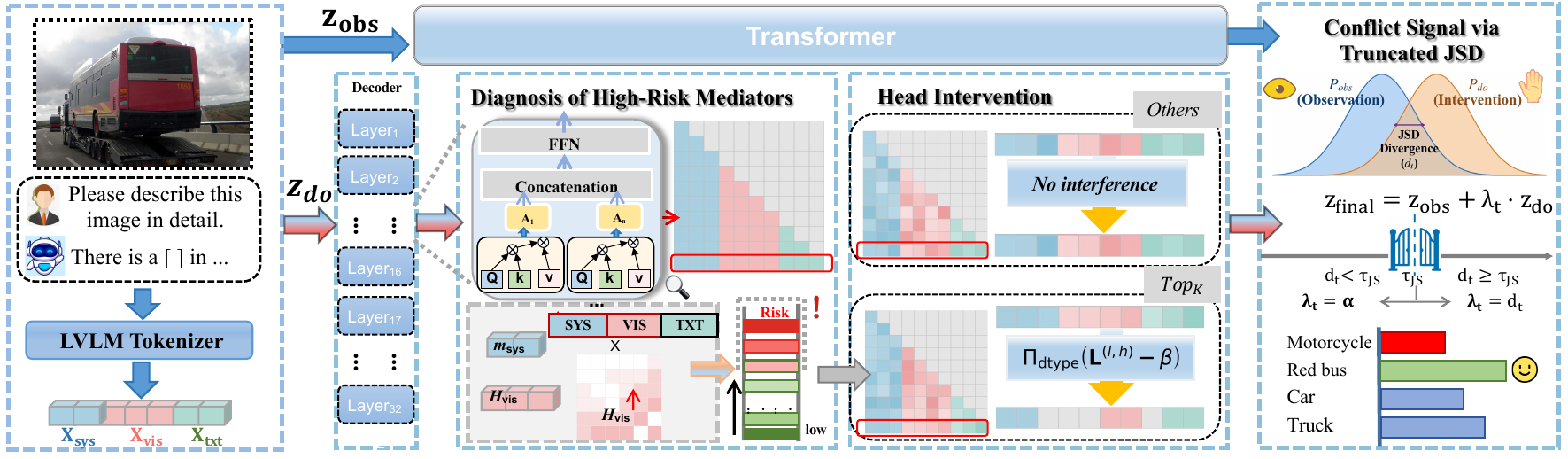}
\caption{
\textbf{Overview of \M.} 
 \textbf{(1) Causal Diagnosis:} Identifying risky mediators $H_R$ by intersecting decision-critical queries with the conjunctive measurement of prior-path activation $m_{\mathrm{sys}}$ and visual uncertainty $H_{\mathrm{vis}}$. 
 \textbf{(2) Causal Intervention:} Executing the $\mathbf{do}$-operator via numerical logit saturation to physically sever pathological shortcuts. 
 \textbf{(3) Adaptive Decoding:} Reconciling observational and interventional distributions via a conflict-gated causal fusion to ensure both faithfulness and linguistic fluency.
}
\label{fig:framework}
\end{figure*}

\subsection{Reliability-Aware Causal Intervention}
\label{sec:intervention}

\textbf{RQ2.}
\textit{How can we physically block the pathological shortcut path $\mathbf{X}_{sys} \to H_R \to Y_t$ without altering fixed model parameters?}

\begin{tcolorbox}[
  tile, size=fbox, boxsep=2mm, boxrule=0pt, top=0pt, bottom=0pt,
  borderline west={1mm}{0pt}{purple!70!pink!30}, 
  colback=yellow!10!white,  
]
\textit{\textbf{\textcolor{gray!50!black}{Insight II}}}:
Effective mitigation requires a surgical intervention that severs the prior-driven shortcut while maintaining the model's structural reasoning capacity. We implement the $\mathbf{do}$-operator via \textit{numerical logit saturation}. By projecting the pre-softmax activations of risky mediators into a low-precision regime, we reset the pathological path to a baseline noise state without inducing the distributional shifts associated with coarse pruning.
\end{tcolorbox}

\noindent\textbf{Causal Intervention via Numerical Saturation.}
As established in \S~\ref{sec:preliminary}, the causal influence of a mediator is physically realized through its attention weights $\mathbf{A} = \text{Softmax}(\mathbf{L})$. To execute the intervention $\mathbf{do}(H_R := \text{noise})$, we modulate the logit-level parents $\mathbf{L}$ directly to achieve finer control over information flow. For any diagnosed risky mediator $(l,h) \in H_R$ at the decision-critical steps $q \in \mathcal{Q}$, we apply a substantial negative bias $\gamma$:
\begin{equation}
    \tilde{\mathbf{L}}^{(l,h)} = 
    \begin{cases}
        \Pi_{\text{dtype}}\left( \mathbf{L}^{(l,h)} - \gamma  \right), & \text{if } (l,h) \in H_R, \\
        \mathbf{L}^{(l,h)}, & \text{otherwise},
    \end{cases}
\end{equation}
where $\gamma$ is a large intervention constant and $\Pi_{\text{dtype}}(\cdot)$ denotes the projection onto the model's numerical precision. 

\noindent\textbf{Causal Rationale of the Intervention.}
While Softmax is shift-invariant in exact arithmetic, its finite-precision implementation provides a unique mechanism for path modulation. By forcing logits into a saturation regime, the intervention achieves two primary objectives:
\begin{enumerate}[itemsep=0pt,topsep=0pt,parsep=0pt,leftmargin=*]
    \item \textit{Suppression of Pathological Shortcuts}: For typical logit ranges, the exponential term $e^{x-\gamma}$ rapidly approaches the machine epsilon $\epsilon_{\text{mach}}$. This precision loss smooths out minor variances encoding spurious language priors, effectively decoupling the mediator from $\mathbf{X}_{sys}$.
    \item \textit{Preservation of Structural Anchors}: Unlike binary masking, this numerical shift allows exceptionally strong structural signals to remain distinct. If an attention peak is sufficiently robust, it can survive the precision collapse, ensuring the intervention does not destroy the essential connectivity of the model's reasoning trajectory.
\end{enumerate}
The resulting distribution $\tilde{\mathbf{A}}^{(l,h)} = \text{Softmax}(\tilde{\mathbf{L}}^{(l,h)})$ represents the post-interventional state where the pathological prior-dependency is neutralized, forcing the model to rely on grounded visual trajectories.

\subsection{Conflict-Gated Cooperative Decoding}
\label{sec:cooperation}

Following the intervention in \S~\ref{sec:intervention}, the model generates two concurrent distributions at each step $t$: the \textit{observational anchor} $P_{obs}$ derived from the full causal graph, and the \textit{interventional candidate} $P_{do}$ derived from $\mathbf{do}(H_R)$. While $P_{do}$ enforces visual grounding, relying on it in isolation may degrade linguistic stability. The final challenge is to adaptively reconcile these distributions to maximize faithfulness.

\textbf{RQ3.}
\textit{How can we dynamically inject interventional evidence to correct hallucinations without disrupting the model's global linguistic stability?}

\begin{tcolorbox}[
  tile, size=fbox, boxsep=2mm, boxrule=0pt, top=0pt, bottom=0pt,
  borderline west={1mm}{0pt}{purple!70!pink!30}, 
  colback=yellow!10!white, 
]
\textit{\textbf{\textcolor{gray!50!black}{Insight III}}}:
Interventional signals should be calibrated by their alignment with the observational manifold. We propose \textit{conflict-gated fusion}: when the two branches reach a consensus, we amplify the interventional signal to solidify evidence; when they diverge, we apply a conservative adjustment to preserve linguistic fluency.
\end{tcolorbox}

\noindent\textbf{Quantifying Causal Conflict via Truncated JSD.}
We measure the disagreement between the biased observation and the debiased intervention via Jensen-Shannon Divergence (JSD). To minimize tail noise, we truncate the vocabulary $\mathcal{V}$ to a candidate set $\mathcal{V}_{t} = \{ y \in \mathcal{V} \mid P_{obs}(y) > \beta \cdot \max_{w} P_{obs}(w) \}$. The conflict signal $d_t$ is:
\begin{equation}
    d_t = \mathrm{JSD}\big(P_{obs}(\cdot|\mathcal{V}_t) \parallel P_{do}(\cdot|\mathcal{V}_t)\big).
\end{equation}
Functionally, $d_t$ quantifies the sensitivity of the prediction to the pathological shortcut. Truncation reduces noise from low-probability tail tokens that fluctuate across branches without changing the selected word.

\noindent\textbf{Conflict-Gated Fusion Strategy.}
Let $\mathbf{z}_{obs}$ and $\mathbf{z}_{do}$ denote the logits from the two branches. We fuse them via a dynamic weight $\lambda_t$ to obtain the final logits $\mathbf{z}_{final} = \mathbf{z}_{obs} + \lambda_t \cdot \mathbf{z}_{do}$. The weight $\lambda_t$ is governed by the conflict magnitude $d_t$ relative to a threshold $\tau_{\mathrm{JS}}$:
\begin{equation}
    \lambda_t = 
    \begin{cases}
        \alpha, & d_t < \tau_{\mathrm{JS}} \\
        d_t, & d_t \ge \tau_{\mathrm{JS}}
    \end{cases}
\end{equation}
This strategy calibrates the interventional influence by its alignment with the observational manifold. In the consensus regime ($d_t < \tau_{\mathrm{JS}}$), we apply a fixed gain $\alpha$ to solidify the grounded evidence. In the conflict regime ($d_t \ge \tau_{\mathrm{JS}}$), the interventional signal acts as a \textit{calibrated correction} where $\lambda_t = d_t$ ensures the shift toward faithfulness remains anchored to the structural stability of $P_{obs}$. The final token $y_t \sim \text{Softmax}(\mathbf{z}_{final})$ thus achieves a principled balance between interventional fidelity and linguistic fluency.

\section{Experiments}
\label{sec:experiments}

\noindent\textbf{Models and Baselines.}
We conduct experiments on three representative LVLMs: \textbf{LLaVA-1.5}~\cite{liu2023visual}, \textbf{Shikra}~\cite{chen2023shikra}, and \textbf{InstructBLIP}~\cite{dai2023instructblip}.
We compare \M~against five inference-time methods: \textbf{ICD}~\cite{ICD}, \textbf{VCD}~\cite{VCD}, \textbf{OPERA}~\cite{huang2024opera}, \textbf{SID}~\cite{SID}, and \textbf{CausalMM}~\cite{CausalMM}.

\noindent\textbf{Benchmarks.}
We assess performance on three standard benchmarks:
\textbf{POPE}~\cite{pope} for object existence verification;
\textbf{CHAIR}~\cite{chair} for hallucination rates in captioning (reporting both CHAIR$_S$ and CHAIR$_I$);
and \textbf{MME}~\cite{mme} for comprehensive perception evaluation (reporting Accuracy and Accuracy+).
Additionally, we employ \textbf{GPT-4V} as a holistic judge to assess open-ended generation quality~\cite{huang2024opera}.

\noindent\textbf{Implementation Details.}
We adopt a unified \emph{Nucleus Sampling} strategy ($p=0.9, T=1.0$) for all models except OPERA (which requires beam search).
For \M, we fix $\alpha{=}2$ and $\beta{=}0.1$, and set $(k,\tau_{\mathrm{JS}})$ to $(0.45, 0.2)$ for LLaVA-1.5, $(0.4, 0.2)$ for InstructBLIP, and $(0.4, 0.2)$ for Shikra.
All experiments are performed on NVIDIA A100 GPUs.
Detailed configurations \textit{cf.} Appendix~\ref{app:implementation}.

\subsection{Main Results}

\paragraph{Results on POPE.}\ 
Table~\ref{tab:pope} reports the performance across three LVLM backbones. \M~achieves consistent improvements, notably reaching an Accuracy of $81.93$\% on LLaVA-1.5 in the \textit{Adversarial} setting.
This gain under high-bias conditions validates that while intensity-based methods (e.g., VCD) fail to decouple visual evidence from linguistic traps, \M~physically severs the pathological shortcut $\mathbf{X}_{sys} \to H_R \to Y_t$ via surgical intervention on risky mediators. This targeted suppression forces the model to rely on stable visual paths rather than learned co-occurrence priors. Unlike OPERA, \M~attains superior defense under standard sampling without search-based overhead, confirming that rectifying structural misalignment is more robust than heuristic-driven amplification. CausalMM achieves competitive POPE by globally adjusting attention via backdoor-based counterfactual reasoning, but its reliance on holistic causal correction without explicitly targeting a sparse set of high-risk attention heads limits its effectiveness when hallucinations are driven by localized structural shortcuts.

\begin{table}[t]
    \centering
    \setlength{\tabcolsep}{4pt}
    \resizebox{\linewidth}{!}{
    \begin{tabular}{c|ccc|ccc|ccc}
        \toprule
        \multirow{2}*{\textbf{Method}} & 
        \multicolumn{3}{c|}{\textbf{\emph{LLAVA-1.5}}} & 
        \multicolumn{3}{c|}{\textbf{\emph{InstructBLIP}}} &
        \multicolumn{3}{c}{\textbf{\emph{Shikra}}} \\
        ~ & Ran$\uparrow$ & Pop$\uparrow$ & Adv$\uparrow$ & Ran$\uparrow$ & Pop$\uparrow$ & Adv$\uparrow$ & Ran$\uparrow$ & Pop$\uparrow$ & Adv$\uparrow$ \\
        \midrule
        Sampling & 85.13 & 82.53 & 76.77 & 86.33 & 80.53 & 77.33 & 81.63 & 80.20 & 75.57 \\
        VCD & 86.84 & 80.87 & 75.33 & 85.97 & 80.63 & 78.77 & 78.13 & 80.60 & 73.80 \\
        ICD & 86.27 & 83.57 & 77.90 & 87.93 & 80.47 & 77.70 & 79.07 & 79.13 & 75.87 \\
        OPERA & 89.27 & \textbf{86.80} & 81.13 & 89.73 & 83.07 & \textbf{81.57} & 83.73 & \textbf{83.27} & \textbf{79.37} \\
        CausalMM & 89.15 & 86.63 & 81.27 & 87.97 & 83.25 & 81.32 & 82.77 & 82.93 & 78.53 \\
        SID & 88.67 & 85.37 & 81.10 & 87.43 & \textbf{83.53} & 78.60 & 82.27 & 79.20 & 79.13 \\
        \hline
        \rowcolor{gray!20}\textbf{Ours} & \textbf{89.33} & 86.70 & \textbf{81.93} & \textbf{89.80} & 83.40 & 80.60 & \textbf{84.97} & 82.27 & 78.77 \\
        \bottomrule
    \end{tabular}
    }
    \caption{Results on POPE. Ran, Pop, and Adv stand for \textit{Random}, \textit{Popular}, and \textit{Adversarial} settings, respectively. The higher score indicates better performance.}
    \label{tab:pope}
\end{table}

\paragraph{Results on CHAIR.}\ 
Table~\ref{tab:chair} reports sentence-level ($C_S$) and instance-level ($C_I$) hallucination rates for long-form captioning. \M~consistently achieves the lowest scores across all backbones, notably reducing $C_I$ to $12.90$ on LLaVA-1.5 and a record $11.98$ on InstructBLIP. These results correspond to a relative $C_I$ reduction of $16.2\%$ and $29.1\%$ over SID, respectively, significantly outperforming strong baselines like OPERA and CausalMM.
These results validate that instance-level hallucinations in descriptive tasks often stem from the pathological propagation of co-occurrence priors. By intervening on risky mediators at decision-critical steps, \M~effectively dismantles these shortcuts, forcing the model to re-verify each entity against visual evidence. The simultaneous reduction in $C_S$ and $C_I$ confirms that our conflict-gated strategy successfully rectifies structural misalignment without sacrificing descriptive richness or linguistic fluency.
\begin{table}[t]
\captionsetup{font=footnotesize}
\centering

\setlength{\tabcolsep}{6pt}  % 核心调大列间距（原2pt，6pt是学术表格舒适间距，不挤也不松散）
\renewcommand{\arraystretch}{0.95}  % 保留原行间距
\fontsize{7pt}{8pt}\selectfont  

{
\begin{tabular}{c|cc|cc|cc} 
\toprule
\multirow{2}{*}{\textbf{Method}} &
\multicolumn{2}{c}{\textbf{\emph{LLAVA-1.5}}} &
\multicolumn{2}{c}{\textbf{\emph{InstructBLIP}}} &
\multicolumn{2}{c}{\textbf{\emph{Shikra}}} \\
& $C_S\downarrow$ & $C_I\downarrow$ & $C_S\downarrow$ & $C_I\downarrow$ & $C_S\downarrow$ & $C_I\downarrow$ \\
\midrule
Sampling            & 57.60 & 17.18 & 56.00 & 16.91 & 56.40 & 15.96 \\
VCD                 & 54.00 & 16.02 & 56.20 & 16.65 & 59.00 & 15.58 \\
ICD                 & 53.40 & 15.67 & 55.40 & 16.52 & 57.60 & 14.68 \\
OPERA               & 49.00 & 13.52 & 45.00 & 13.15 & 53.60 & 14.20 \\
CausalMM            & 47.60  & 13.49 & 43.20 & 13.05 & 53.00 & 14.27 \\
SID                 & 53.40 & 15.40 & 54.00 & 16.89 & 54.20 & 14.79 \\
\hline
\rowcolor{gray!20}\textbf{Ours}       & \textbf{46.40} & \textbf{12.90} & \textbf{42.40} & \textbf{11.98} & \textbf{52.00} & \textbf{13.66} \\
\bottomrule
\end{tabular}
}
\caption{Results on the CHAIR. $C_S$ and $C_I$ denote $\text{CHAIR}_S$ and $\text{CHAIR}_I$ (the smaller score indicates fewer hallucinations).}
\label{tab:chair}
\end{table}
% \begin{table}[htbp]
% \centering
% \caption{\small LLaVA-Bench (In-the-Wild) evaluation results.}
% \vspace{-0.35em}
% \resizebox{1.0 \linewidth}{!}{
% \begin{tabular}{l|*{5}{>{\centering\arraybackslash}p{1.45cm}}}
%  Methods & Convs $\uparrow$ & Detail $\uparrow$ & Complex\;$\uparrow$ & All $\uparrow$ & Average $\uparrow$\\ 
% \midrule
% LLaVA-1.5 & -- \basex{0.0} & {--} \basex{0.0} & -- \basex{0.0} & {--} \basex{--} & {--} \basex{0.0} \\
% \ + OPERA & {--} \up{--} & -- \downbad{--} & \sethlcolor{lightpink} \hl{--}
% \up{--} & {--} \downbad{--}  & {--} \downbad{--} \\
% \ + ICD & -- \downbad{--} & -- \downbad{--} & -- \downbad{--}
% & -- \downbad{--} & -- \downbad{--} \\
% \ + VCD & -- \downbad{--} & -- \downbad{--} & -- \downbad{--}
% & -- \downbad{--} & -- \downbad{--} \\
% \ + SID & -- \downbad{--} & {--} \downbad{--} & {--} \up{--} & -- \downbad{--}  & -- \downbad{--} \\\midrule
% \ {+ Our} & \sethlcolor{lightpink}\hl{--} \up{--} & \sethlcolor{lightpink}\hl{--} \up{--} & {--} \up{--} & \sethlcolor{lightpink}\hl{--} \up{--} & \sethlcolor{lightpink}\hl{--} \up{--} \\
% \end{tabular}}\label{tab:llavabench} 
% \end{table}

\begin{figure}[ht]
\centering
\includegraphics[width=0.48\textwidth]{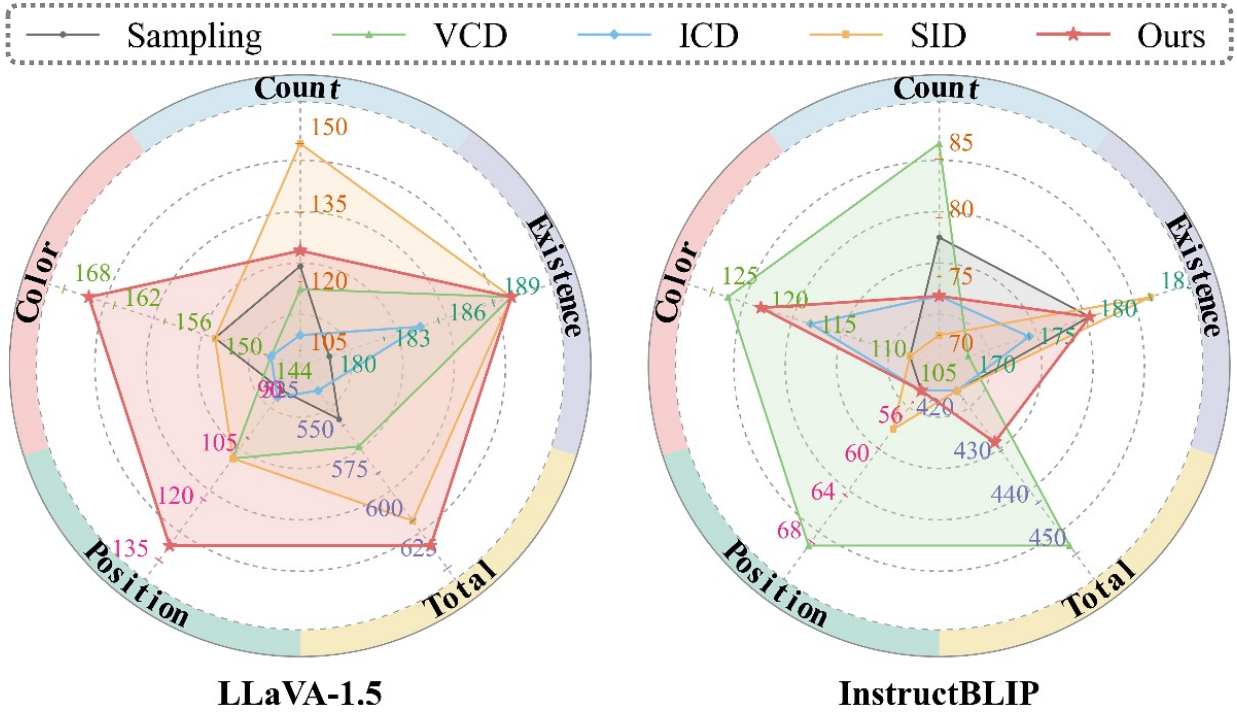}
\caption{
Performance on the MME benchmark. Higher scores indicate better effectiveness. \M~achieves the highest total scores across all evaluated backbones, particularly excelling in evidence-driven subsets Position and Color.}
\vspace{-0.5em}
\label{fig:MME}
\end{figure}

\begin{figure}[t]
\centering
\includegraphics[width=0.9\linewidth]{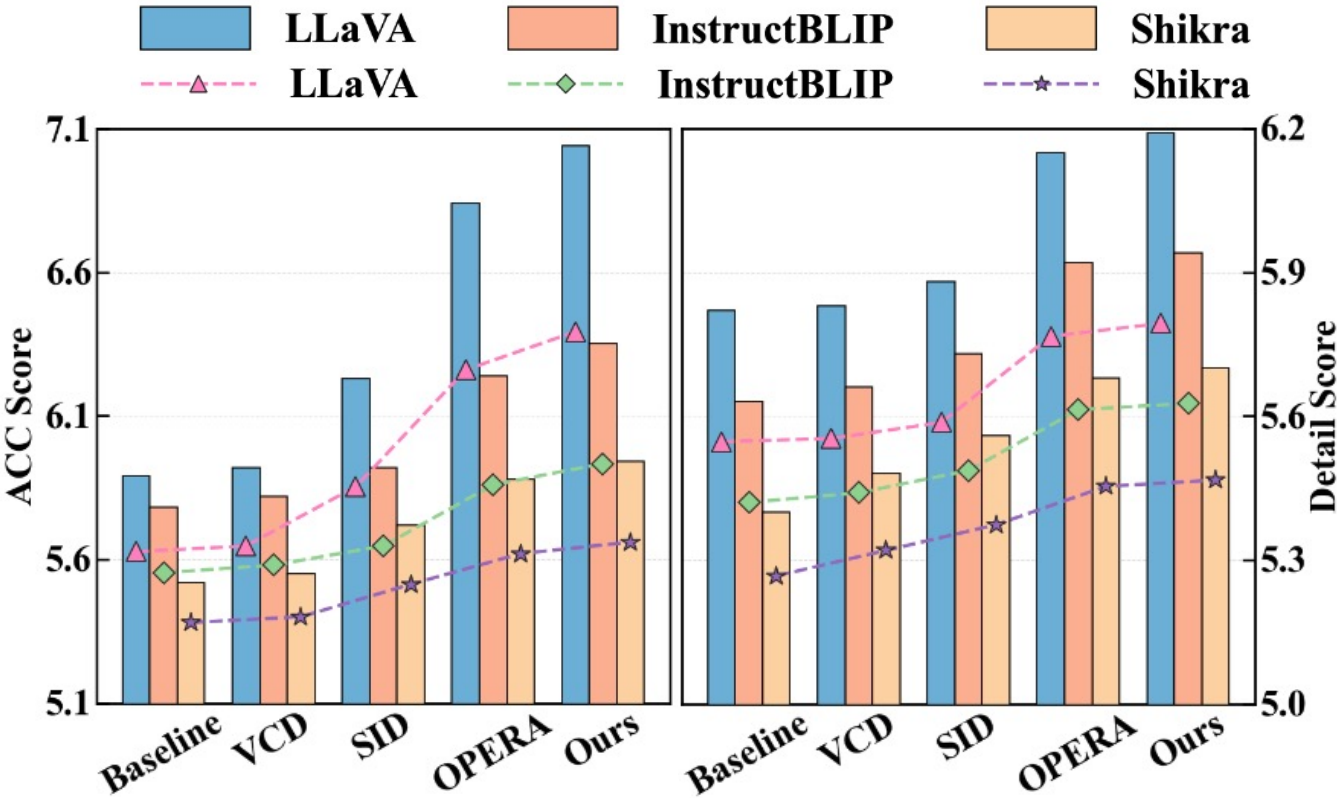}
\caption{
GPT-4V-assisted hallucination evaluation. Left: Correctness (higher = less hallucination); Right: Detailedness. Lines highlight within-backbone trends.
}
\label{fig:gptv4}
\end{figure}

\begin{figure}[t]
  \centering
  \includegraphics[width=0.9\linewidth]{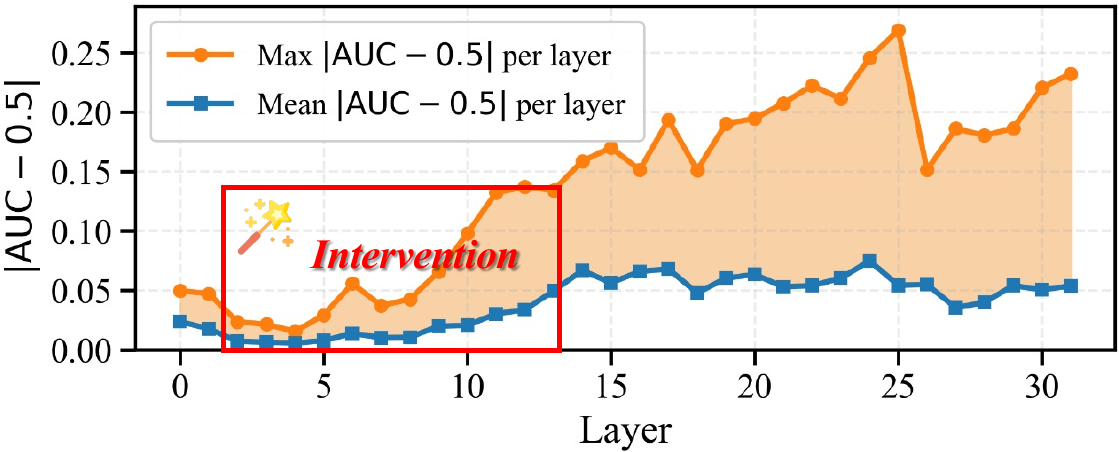}
  \caption{
  \textbf{Layer-wise diagnostic strength of visual uncertainty.} The $|\mathrm{AUC}-0.5|$ metric (Y-axis) quantifies the separability between faithful and hallucinated samples based on entropy probes. While deeper layers exhibit higher peak discriminative strength, the red box indicates the optimal \textit{intervention window} (early-to-mid layers) where structural rectification yields the most significant grounding improvements.
  }
  \label{fig:ablation-layer}
\end{figure}

\paragraph{Results on MME Benchmark.}\ 
Fig.~\ref{fig:MME} illustrates the performance across fine-grained subsets. \M~consistently improves the total score across all backbones (results for Shikra \textit{cf.} Appendix~\ref{app:mme}), notably increasing LLaVA-1.5's score to 613.33, surpassing the strong baseline SID (600.00). The most significant gains occur in evidence-dependent dimensions, such as \textit{Position} ($93.33 \to 131.37$) and \textit{Color} ($150.00 \to 165.00$).
These results provide empirical proof that attribute-level assertions are highly susceptible to prior-path dominance. By disrupting the pathological shortcut at decision-critical steps, \M~forces the model to re-verify fine-grained visual evidence, explaining the marked improvements in the \textit{Position} and \textit{Color} subsets. The consistent enhancement across backbones confirms that rectifying structural misalignment ensures model assertions remain anchored in visual reality rather than linguistic plausibility.

\paragraph{Results on GPT-4V Evaluation.}\ 
Fig.~\ref{fig:gptv4} summarizes the qualitative evaluation using GPT-4V as a holistic judge, following a standardized evaluation protocol in Appendix~\ref{app:gpt4v}. Across all backbones, \M~consistently improves \textit{Correctness} while simultaneously enhancing \textit{Detailedness}, indicating a superior fidelity--detail trade-off. Specifically, on LLaVA-1.5, \M~raises Correctness from $5.89$ to $7.04$ and Detailedness from $5.82$ to $6.19$. These upward trends confirm that our gains are not achieved via overly conservative or ``evasive'' descriptions, but via authentic visual grounding.
These results further validate our causal intervention mechanism. Unlike global re-weighting strategies that often lead to distribution shifts or information loss, \M~targets risky mediators only at decision-critical steps. By surgically severing pathological shortcuts while preserving the structural integrity of the generative manifold, our method effectively suppresses prior-driven behaviors without incurring a loss in descriptive richness. Consequently, \M~rectifies the dynamic structural misalignment of baseline decoding, ensuring responses are both faithful and expressive.

\subsection{Ablation Study}

\begin{figure}[t]
 \centering
 \includegraphics[width=0.9\linewidth]{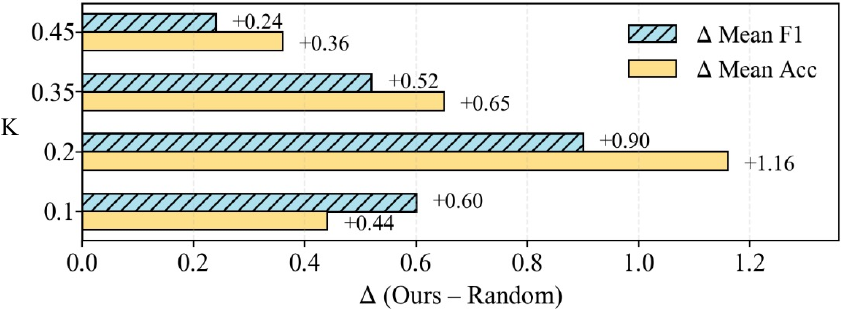}
 \caption{
 \textbf{Effectiveness of diagnosis-driven head selection.} We report the performance gains of \M~over a random-intervention baseline across different intervention ratios $K$ on the POPE benchmark. Improvements in Mean Accuracy ($\Delta$ Mean Acc) and Mean F1 ($\Delta$ Mean F1) consistently validate that targeting specific \textit{risky mediators} is superior to stochastic head suppression.
 }
 % \vspace{-3mm}
 \label{fig:top-k}
\end{figure}

\begin{figure}[t]
  \centering
  \includegraphics[width=1\linewidth]{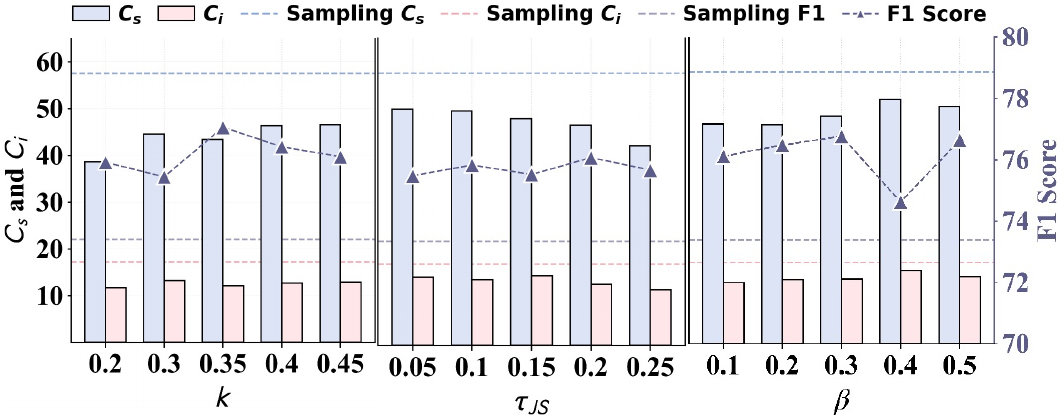}
  \caption{Impact of $k$, $\tau_{\mathrm{JS}}$, and $\beta$ on hallucination and informativeness in LLaVA-1.5, evaluated on $500$ COCO samples.
  }
  \label{fig:parameter}
\end{figure}

\begin{figure*}[ht]
  \centering
  \includegraphics[width=0.9\linewidth]{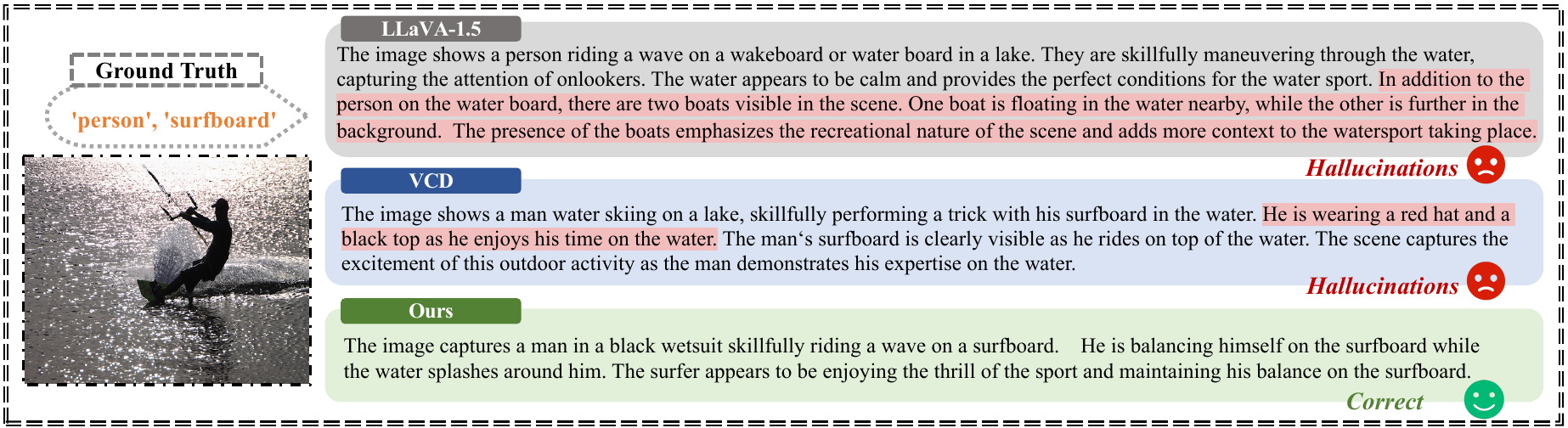}
\caption{Open-ended captioning comparison. Hallucinations are marked in red. \M~effectively mitigates hallucination while preserving descriptive richness.}
    \label{fig:llava-Illustration}
    \vspace{-3mm}
\end{figure*}

\begin{figure}[ht]
  \centering
  \includegraphics[width=0.95\linewidth]{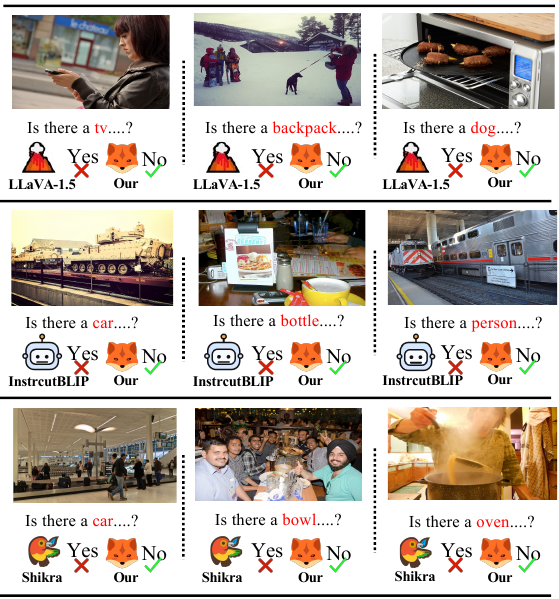}
  \caption{Qualitative examples of VLMs on POPE for object existence prediction. Red: Hallucination; Green: Correct predictions.}
  \label{fig:pope}
  \vspace{-3mm}
\end{figure}

\paragraph{Validation of Diagnosis and Structural Intervention.}\ 
As shown in Fig.~\ref{fig:ablation-layer}, visual uncertainty signals exhibit a non-uniform distribution across layers. Although peak discriminative patterns (Max $|\mathrm{AUC}-0.5|$) are prominent in deeper layers, we empirically find that structural intervention at these late stages yields diminishing returns for hallucination mitigation. In contrast, targeting early-to-mid layers (Layers 2--13) more effectively regulates the evidence aggregation process, achieving a superior trade-off between faithfulness and fluency. Consequently, we define this critical window as our default intervention range.

To verify the precision of our head localization, we compare  diagnosis-driven selection against a \textit{random selection} baseline under a constant budget $K$. As illustrated in Fig.~\ref{fig:top-k}, \M~outperforms random perturbations across $K$ values, confirming that the observed gains stem from accurately isolating a sparse set of \textit{risky mediators} rather than stochastic noise. To further eliminate potential confounders from sampling stochasticity, we evaluate \M~in a deterministic setting (i.e., greedy decoding). In this regime, \M~maintains clear advantages over strong baselines such as VCD and SID (details \textit{cf.} Appendix~\ref{app:Greedy_Decoding}). These results reinforce that our intervention effectively addresses the dynamic structural misalignment by surgically severing pathological shortcuts.

\paragraph{Hyperparameter Sensitivity.}\ 
Fig.~\ref{fig:parameter} investigates the sensitivity of \M~using $500$ COCO samples with LLaVA-1.5, varying the per-layer head suppression ratio $k$, the conflict threshold $\tau_{\mathrm{JS}}$, and the truncation ratio $\beta$. Overall, \M~remains stable across a broad range of values, with $C_S$, $C_I$, and F1 changing smoothly, suggesting that the framework does not rely on fragile tuning.
Among these, $\tau_{\mathrm{JS}}$ most strongly governs the trade-off between faithfulness and informativeness. A larger $\tau_{\mathrm{JS}}$ makes the gate more selective, suppressing prior-driven \textit{shortcut propagation} more aggressively and typically reducing hallucination rates; however, excessive thresholds may over-constrain the generation and slightly decrease F1. In contrast, $k$ exhibits milder effects: increasing $k$ consistently lowers hallucination with minimal F1 fluctuations. Finally, $\beta$ controls candidate truncation in conflict estimation, where moderate values yield the most robust performance. Similar trends are observed on InstructBLIP and Shikra (More details \textit{cf.} Appendix~\ref{app:add_ablation}).

\subsection{Case Study}

Qualitative analysis confirms \M's precision.
Fig.~\ref{fig:llava-Illustration} shows that in open-ended captioning, \M~restores visual grounding by suppressing prior-driven behaviors. The resulting descriptions remain strictly faithful to visual evidence without incurring a loss in descriptive richness, effectively addressing the dynamic structural misalignment inherent in standard autoregressive decoding. These examples also show that the intervention is token-selective: it does not erase the full descriptive context, but suppresses unsupported entities that emerge from prior-dominated mediators. Similarly, as illustrated in Fig.~\ref{fig:pope}, while baselines succumb to existence biases by fabricating non-existent objects, \M~rectifies these errors by intervening on risky mediators to dismantle pathological shortcuts. This behavior suggests localized causal repair rather than generic conservativeness.

\section{Conclusion \& Future Work}
\label{sec:conclusion}

In this paper, we challenge the prevailing attention intensity assumption and propose \M, a training-free causal framework for mitigating hallucination by rectifying dynamic structural misalignment. By treating attention heads as dynamic mediators, \M~uses visual attention entropy to identify and suppress risky mediators that form pathological shortcuts at decision-critical steps. A conflict-gated cooperative decoding strategy further balances visual faithfulness and linguistic fluency. Extensive experiments show that \M~achieves superior hallucination reduction while preserving generation quality and maintaining the latency regime of contrastive decoding baselines.
Future work will extend this causal diagnosis to compositional failures, including relation, attribute, and counting errors, and to video-language and tool-augmented LVLMs, where temporal or external evidence may introduce new shortcuts. We will also study adaptive mediator discovery under distribution shift.

\clearpage
\newpage
\section*{Acknowledgements}
This work was supported by the Sichuan Provincial Natural Science Foundation Project (Grant No. 2026NSFSC0427), the Chengdu Technological Innovation and R\&D Project (Grant No. 2026YF0800348GX), and the China Scholarship Council (CSC, Grant No. 202506070076).

\section*{Impact Statement}
This paper aims to improve the faithfulness of large vision-language models by reducing object hallucinations at inference time. More reliable multimodal generation can benefit applications such as assistive technologies, educational tools, content understanding, and information retrieval, where users may depend on model outputs to interpret visual evidence. By framing hallucination as a causal shortcut induced by dynamic mediator heads, the proposed method also provides a more interpretable perspective on when and how visual grounding fails.

At the same time, our method should not be viewed as a guarantee of factual correctness. LVLMs may still inherit dataset biases, fail under distribution shift, or produce unsupported claims in complex scenes beyond the object-centric settings studied here. Reducing hallucination may also increase users' trust in model outputs, so deployment in safety-critical or high-stakes domains should retain human oversight, uncertainty communication, and task-specific evaluation. We do not anticipate specific negative societal impacts beyond those generally associated with multimodal foundation models, and we hope this work encourages more transparent and accountable LVLM deployment.

% Acknowledgements should only appear in the accepted version.

\bibliographystyle{icml2026}
\bibliography{example_paper}

\newpage
\appendix
\onecolumn
\section{Additional Evidence for Motivation and Uncertain Signal}
\label{app:motivation}

\subsection{Statistical Setup and Window Definitions}
\label{app:stat_setup}

\textbf{Dataset and Sampling.}
We conduct a fine-grained head-level statistical analysis on LLaVA-1.5 to provide empirical grounding for our causal assumptions. We utilize two balanced subsets: $1{,}000$ faithful samples and $1{,}000$ hallucinated samples. For each sample, we extract the raw attention probability matrices during autoregressive generation to characterize the internal information flow. We then compute metrics across the layer--head dimension to localize structural misalignment during decision-critical multimodal integration.

\textbf{Window Definitions.}
To verify that our identified diagnostic signals are not transient artifacts but reflect robust structural properties of the causal graph, we define two query-position windows denoted by the length \texttt{tail}. These windows are defined on the \emph{Text-Tail} query set $\mathcal{Q}_{\text{tail}}$, representing post-image textual positions where language priors typically begin to compete with visual evidence. The \emph{Prefill-Last} decision point, serving as the multimodal handshake, is analyzed separately as the trajectory initialization bottleneck.

\begin{itemize}
    \item \textbf{Instantaneous Window ($\text{tail}=1$):} 
    Statistics are computed using only the final query position of the current step. This setting captures the mediator behavior at the precise \emph{instant of causal decision}, directly aligning with the ``Decision-Critical Steps'' identified in our SCM. It is uniquely sensitive to the transient structural ``locking'' where a head decouples from visual evidence to favor priors.

    \item \textbf{Smoothed Window ($\text{tail}=32$):} 
    Statistics are aggregated over the 32 most recent text query positions. This setting filters out token-level stochasticity to reveal the \emph{persistent structural bias} of specific attention heads across a broader local context.
\end{itemize}

\textbf{Causal Complementarity.}
These windows provide multi-scale validation of our framework: $\text{tail}=1$ identifies the immediate trigger of pathological shortcuts at decision boundaries, while $\text{tail}=32$ confirms the stability of risky mediators as unreliable information conduits. Reporting both settings demonstrates that while signal magnitude may vary, the underlying structural characteristics of \emph{risky mediators}—specifically their tendency to mediate prior-driven influence—remain invariant to temporal smoothing.

\subsection{Global Differences in Visual Attention}
\label{app:global_diff}

We investigate whether object hallucination originates from a \emph{quantitative} deficit in global visual grounding, as suggested by the prevailing attention intensity assumption. By analyzing the sample-level distribution of visual attention mass, we provide empirical evidence that hallucination is a structural pathology rather than a simple signal-magnitude failure.

\textbf{Head-level Visual Mass.}
For each layer $l$ and head $h$, let $\mathbf{L}^{(l,h)}$ denote the pre-softmax attention activations and
\begin{equation}
    \mathbf{A}^{(l,h)}=\mathrm{Softmax}\!\left(\mathbf{L}^{(l,h)}\right)
\end{equation}
denote the resulting attention weights. Given the visual token indices $\mathcal{I}_{vis}=[v_s, v_e)$ and the windowed query set $\mathcal{Q}_{\text{tail}}$ (Appendix~\ref{app:stat_setup}), the head-level visual attention mass is defined as:
\begin{equation}
    m^{(l,h)}_{V,\text{tail}}
    =
    \frac{1}{|\mathcal{Q}_{\text{tail}}|}
    \sum_{q \in \mathcal{Q}_{\text{tail}}}
    \sum_{k \in \mathcal{I}_{vis}}
    \mathbf{A}^{(l,h)}_{q,k}
    \in [0,1].
\end{equation}
This quantity represents the total causal weight a specific mediator allocates to the visual path $X_{vis} \to H$ during the generation window.

\textbf{Sample-level Aggregation.}
To evaluate the model's global observational state, we aggregate the mass across all latent mediators:
\begin{equation}
    m_{V,\text{tail}}
    =
    \frac{1}{L \times H}
    \sum_{l=1}^{L}
    \sum_{h=1}^{H}
    m^{(l,h)}_{V,\text{tail}}.
\end{equation}

\begin{figure}[t]
    \centering
    \includegraphics[width=1\linewidth]{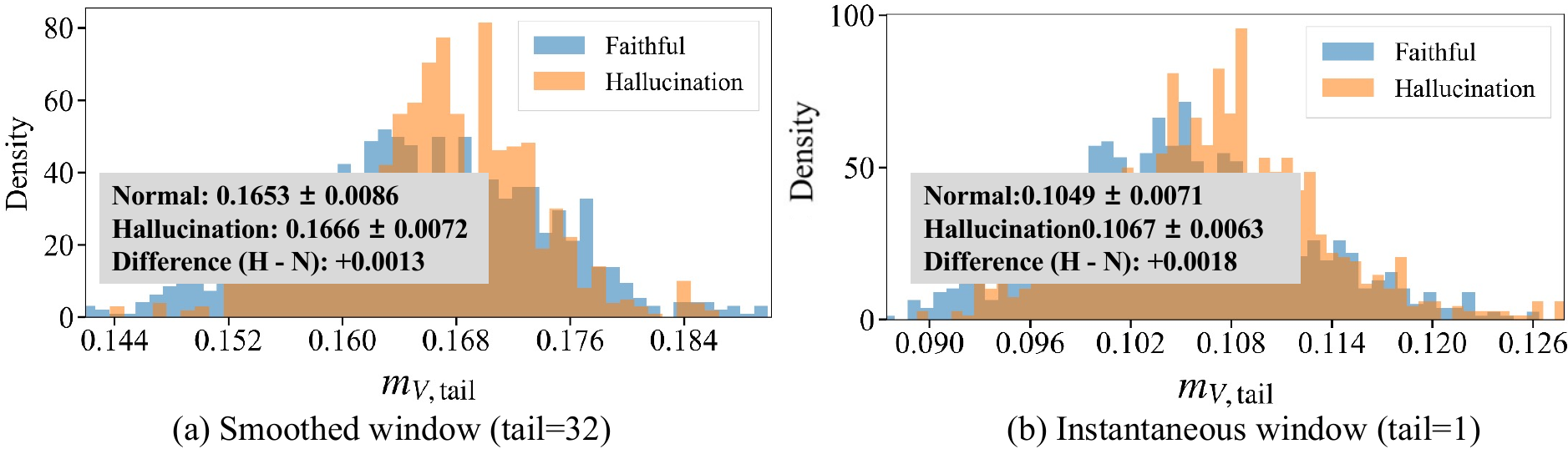}
    \caption{
    \textbf{Hallucination is not explained by a global reduction in visual attention mass.}
    Comparing the distribution of the sample-level global visual attention mass $m_{V,\mathrm{tail}}$ (averaged over all layers and heads) between 1,000 Faithful (blue) and 1,000 Hallucinated (orange) samples on LLaVA-1.5-7B.
    \textbf{(a)} Smoothed window ($\mathrm{tail}=32$) aggregating statistics over recent post-image text queries.
    \textbf{(b)} Instantaneous window ($\mathrm{tail}=1$) measuring the decision-step query.
    The two distributions largely overlap under both windows, ruling out a coarse global-intensity account and motivating head-level structural diagnosis.
    }
    \label{fig:app_visual_mass}
\end{figure}

\noindent\textbf{Counter-evidence for Intensity-based Explanations.}
We compare the distributions of $m_{V,\text{tail}}$ for faithful and hallucinated samples in Figure~\ref{fig:app_visual_mass}. Critically, hallucinated outputs do not exhibit a statistically significant reduction in global visual attention mass. As illustrated in Figure~\ref{fig:app_visual_mass}(a--b), the distributions for both groups overlap substantially across both instantaneous ($\text{tail}=1$) and smoothed ($\text{tail}=32$) windows. 

These results provide a formal refutation of the coarse \emph{global-intensity} hypothesis. The evidence confirms that the model maintains sufficient global visual engagement, yet fails due to \emph{structural misalignment} at decision-critical steps. This supports our shift from global magnitude boosting toward surgical intervention on sparse, high-risk mediators that facilitate pathological shortcuts despite nominal global attention.

\subsection{Relationship between System Prompts and Hallucinations}
\label{app:sys_hallucination}

This section provides additional evidence that reliance on system/prefix tokens is a structured signal rather than random noise, and exhibits non-trivial separability between Faithful and Hallucinated samples on our analysis set. We report results under both $\text{tail}=1$ and $\text{tail}=32$ windows defined in Appendix~\ref{app:stat_setup}, with hallucination treated as the positive class.

\paragraph{System-reliance metric.}
For each layer $l$ and head $h$, let $\mathbf{A}^{(l,h)}=\mathrm{Softmax}(\mathbf{L}^{(l,h)})$ denote attention weights. Let the system/prefix indices be $\mathcal{I}_{sys}=[0,s_{img})$, where $s_{img}$ is the starting index of visual tokens. Given the windowed query set $\mathcal{Q}_{\text{tail}}$, we define the system-attention mass:
\begin{equation}
    m^{(l,h)}_{sys,\text{tail}}
    =
    \frac{1}{|\mathcal{Q}_{\text{tail}}|}
    \sum_{q \in \mathcal{Q}_{\text{tail}}}
    \sum_{k \in \mathcal{I}_{sys}}
    \mathbf{A}^{(l,h)}_{q,k}
    \in [0,1].
\end{equation}
A larger $m^{(l,h)}_{sys,\text{tail}}$ indicates stronger reliance on system/prefix priors within the observed window.

\paragraph{Head-wise discriminability.}
We compute the ROC-AUC for each head $(l,h)$ using $m^{(l,h)}_{sys,\text{tail}}$ as the score and report its centered value:
\begin{equation}
    \Delta \mathrm{AUC}^{(l,h)}=\mathrm{AUC}^{(l,h)}-0.5.
\end{equation}
Here, $\Delta \mathrm{AUC}^{(l,h)}>0$ indicates that higher system reliance correlates with hallucination, while $\Delta \mathrm{AUC}^{(l,h)}<0$ indicates the opposite tendency.

\paragraph{Weighted sample-level aggregation (analysis-only).}
To summarize the system-reliance signal into a sample-level score for visualization, we aggregate head-wise masses with $\Delta\mathrm{AUC}$ weighting:
\begin{equation}
    s_n(m_{sys,\text{tail}})
    =
    \frac{
    \sum_{l,h} m^{(l,h)}_{sys,\text{tail}}(n)\cdot \Delta \mathrm{AUC}^{(l,h)}
    }{
    \sum_{l,h} \left|\Delta \mathrm{AUC}^{(l,h)}\right|
    }.
\end{equation}
This aggregation is used \emph{only} as a diagnostic probe on the analysis set; it is not a deployed hallucination detector.

\begin{figure}[t]
    \centering
    \includegraphics[width=0.65\linewidth]{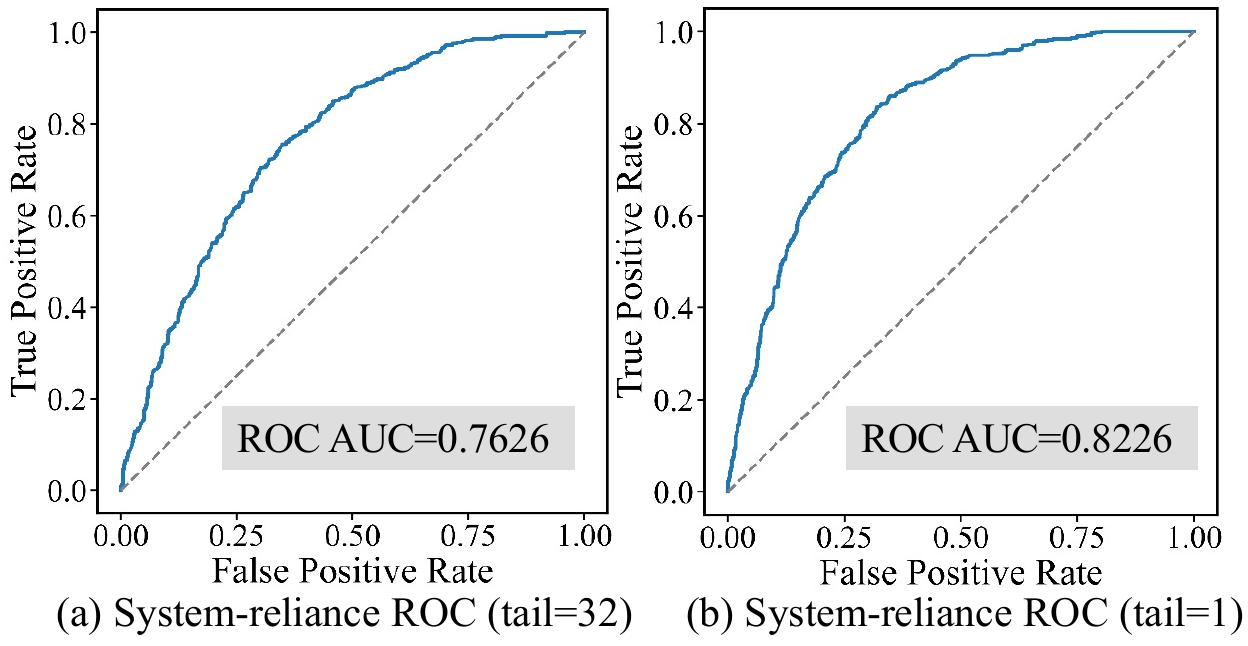}
    \caption{
    \textbf{All-head aggregated system reliance separates hallucinated from faithful samples.}
    We compute an analysis-only sample-level score by aggregating head-wise system attention mass $m^{(l,h)}_{sys,\text{tail}}$ over \emph{all} heads with $\Delta \mathrm{AUC}^{(l,h)}$ weighting (Appendix~\ref{app:sys_hallucination}); a higher score indicates stronger reliance on system/prefix tokens.
    \textbf{(a)} Smoothed window ($\text{tail}=32$), AUC=0.7626.
    \textbf{(b)} Instantaneous window ($\text{tail}=1$), AUC=0.8226.
    The stronger separability under $\text{tail}=1$ indicates that system reliance peaks at instantaneous decision steps.
    }
    \label{fig:app_sys_roc}
\end{figure}

\paragraph{Results and analysis.}
Figure~\ref{fig:app_sys_roc} plots ROC curves based on $\{s_n(m_{sys,\text{tail}})\}$.
The instantaneous window ($\text{tail}=1$) yields an AUC of 0.8226, while the smoothed window ($\text{tail}=32$) yields an AUC of 0.7626, indicating a stable and non-trivial signal.
The higher AUC under $\text{tail}=1$ suggests that system reliance is most pronounced at the instantaneous decision step.
In our method, this system-reliance cue is combined with visual uncertainty to form the joint risk score used for mediator localization in the main text.

\subsection{Visual Uncertainty and Joint Risk Scoring}
\label{app:visual_uncertainty}

This section provides complementary evidence for the joint risk formulation used in the main text (Section \ref{sec:diagnosis}) to localize high-risk mediators.
Rather than treating visual uncertainty as an isolated cue, we study how \emph{visual-side instability} and \emph{system-prior reliance} co-occur at the head level, and show that their multiplicative interaction yields a \emph{sparse} and \emph{stable} diagnostic signal for hallucination risk.
All statistics follow the same windowed query set $\mathcal{Q}_{\text{tail}}$ and token partitions introduced earlier (Appendix~\ref{app:stat_setup}--\ref{app:sys_hallucination}), with hallucination treated as the positive class.

\paragraph{Visual uncertainty via modality-specific entropy.}
Given attention weights $\mathbf{A}^{(l,h)}$, we quantify visual-side uncertainty by the dispersion of attention \emph{within the visual subspace}. For each query position $q\in\mathcal{Q}_{\text{tail}}$, we re-normalize attention over visual keys $\mathcal{I}_{vis}$:
\begin{equation}
    p^{(l,h)}_{q,k}
    =
    \frac{\mathbf{A}^{(l,h)}_{q,k}}
    {\sum_{j \in \mathcal{I}_{vis}} \mathbf{A}^{(l,h)}_{q,j} + \varepsilon},
    \quad k \in \mathcal{I}_{vis},
\end{equation}
where $\varepsilon$ is a small constant for numerical stability. We then define the modality-specific visual entropy:
\begin{equation}
    H^{(l,h)}_{vis}(q)
    =
    -\sum_{k \in \mathcal{I}_{vis}}
    p^{(l,h)}_{q,k}
    \log\!\left(p^{(l,h)}_{q,k} + \varepsilon\right),
\end{equation}
and average it over the tail window:
\begin{equation}
    H^{(l,h)}_{vis,\text{tail}}
    =
    \frac{1}{|\mathcal{Q}_{\text{tail}}|}
    \sum_{q \in \mathcal{Q}_{\text{tail}}}
    H^{(l,h)}_{vis}(q).
\end{equation}
A larger $H^{(l,h)}_{vis,\text{tail}}$ indicates more dispersed (less certain) routing of visual evidence at decision-critical steps.

\paragraph{Joint risk score for mediator localization.}
Visual uncertainty alone is insufficient to characterize hallucination: a head can be visually uncertain yet benign if it does not propagate language priors into generation. Following the main text, we define a head-level joint risk score as the interaction between visual uncertainty and system reliance:
\begin{equation}
    S^{(l,h)}_{\text{tail}}
    =
    m^{(l,h)}_{sys,\text{tail}}
    \cdot
    H^{(l,h)}_{vis,\text{tail}}.
\end{equation}
The multiplicative form suppresses benign cases where only one factor is high and highlights heads exhibiting the co-activation pattern most consistent with prior-driven shortcut behavior.

\paragraph{TopHeads-projected joint risk as a sample-level diagnostic probe.}
To summarize head-wise joint risk into a sample-level score for analysis, we compute a directional weight for each head based on its head-wise ROC-AUC:
\begin{equation}
    \Delta \mathrm{AUC}^{(l,h)}=\mathrm{AUC}^{(l,h)}-0.5,
\end{equation}
rank heads by $|\Delta \mathrm{AUC}^{(l,h)}|$, and denote the Top-$K$ set by $\mathcal{H}_K$.
For each sample $n$, we compute the weighted projection
\begin{equation}
    s_n(S_{\text{tail}})
    =
    \frac{
    \sum_{(l,h)\in\mathcal{H}_K}
    \widehat{S}^{(l,h)}_{\text{tail}}(n)\cdot \Delta \mathrm{AUC}^{(l,h)}
    }{
    \sum_{(l,h)\in\mathcal{H}_K}
    |\Delta \mathrm{AUC}^{(l,h)}|
    },
\end{equation}
where $\widehat{S}^{(l,h)}_{\text{tail}}(n)$ is a per-head z-score computed using dataset-level statistics (mean and standard deviation) for that head.
This TopHeads projection is used \emph{only} for diagnosing sparsity/stability on the analysis set; it does not prescribe the execution-time intervention, which follows a per-layer proportional budget in the main method.

\begin{figure}[ht]
    \centering
    \includegraphics[width=1\linewidth]{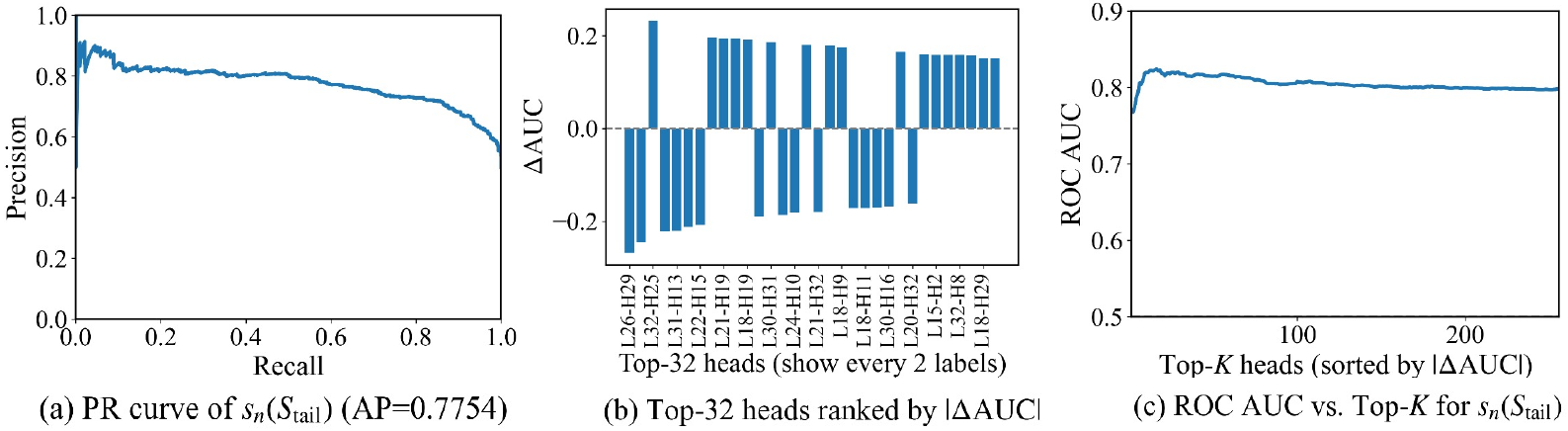}
    \caption{
    \textbf{Sample-level diagnosis and sparsity analysis of the joint risk score.}
    All statistics are computed on the \textit{Text-Tail} query set with an instantaneous window ($\text{tail}=1$).
    We define the head-level joint risk as $S^{(l,h)}_{\text{tail}} = m^{(l,h)}_{sys,\text{tail}}\cdot H^{(l,h)}_{vis,\text{tail}}$ and form a sample-level diagnostic score $s_n(S_{\text{tail}})$ by projecting standardized head-wise scores onto the Top-$K$ heads ranked by $|\Delta\mathrm{AUC}^{(l,h)}|$.
    \textbf{(a)} Precision--Recall curve of the Top-32 projected joint-risk score.
    \textbf{(b)} The Top-32 heads ranked by $\Delta\mathrm{AUC}^{(l,h)}$, illustrating that the signal concentrates on a sparse subset of mediators.
    \textbf{(c)} ROC AUC as a function of $K$, showing rapid saturation with increasing $K$.
    }
    \label{fig:app_joint_stability}
\end{figure}

\paragraph{Sparsity and a global-aggregation reference point.}
Figure~\ref{fig:app_joint_stability} shows that the discriminative power of $s_n(S_{\text{tail}})$ saturates quickly as $K$ increases, indicating that the diagnostic signal is concentrated in a small subset of heads.
As a reference point, if we aggregate $S^{(l,h)}_{\text{tail}}$ over \emph{all} heads without Top-$K$ selection (i.e., replacing $\mathcal{H}_K$ by all $(l,h)$), the ROC AUC is $0.7852$, which is lower than the Top-32 projection used in the main text (ROC AUC $0.8180$). This suggests that global aggregation tends to dilute the localized signal rather than amplify it.

\subsection{Visual Evidence of Uncertainty and Structural Reconfiguration Post-Intervention}
\label{app:intervention_visual}

In this section, we provide head-level visual evidence to support the \emph{Risky Mediator} localization in the main text.
We select representative heads ranked highly by the joint risk score $S^{(l,h)}_{\text{tail}}$ (Appendix~\ref{app:visual_uncertainty}) and visualize how their attention structure changes before and after our logit-level intervention.
For each head, we fix a representative sample and a decision-critical query step, and compare the Baseline (left) against the Intervention (right).

\begin{figure*}[!t]
\centering
\setlength{\tabcolsep}{0.5em}
\renewcommand{\arraystretch}{0.8}
\begin{tabular}{c}
\subfloat[Representative risky head (example 1).]{%
  \includegraphics[width=0.97\linewidth]{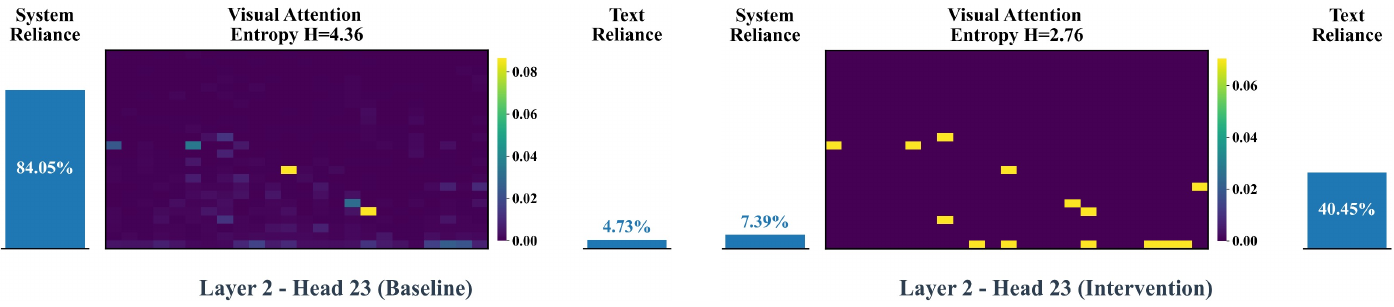}
}\\[0.45em]
\subfloat[Representative risky head (example 2).]{%
  \includegraphics[width=0.97\linewidth]{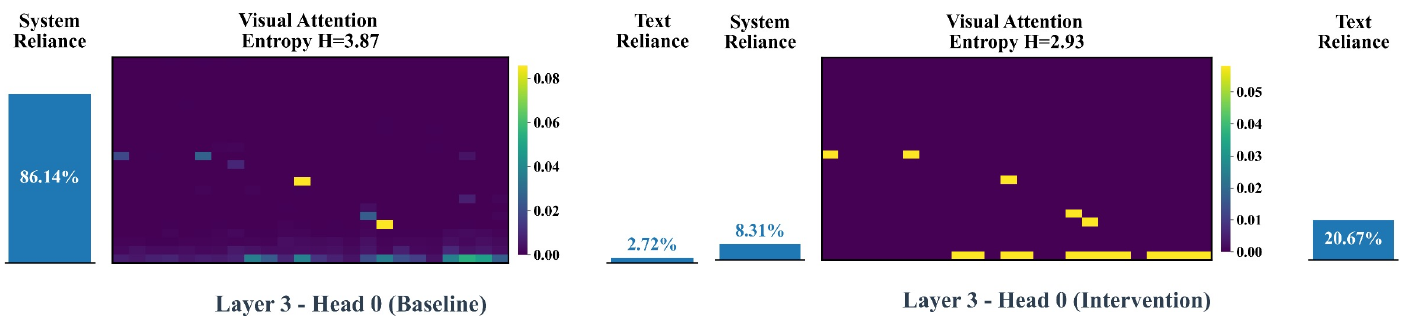}
}\\[0.45em]
\subfloat[Representative risky head (example 3).]{%
  \includegraphics[width=0.97\linewidth]{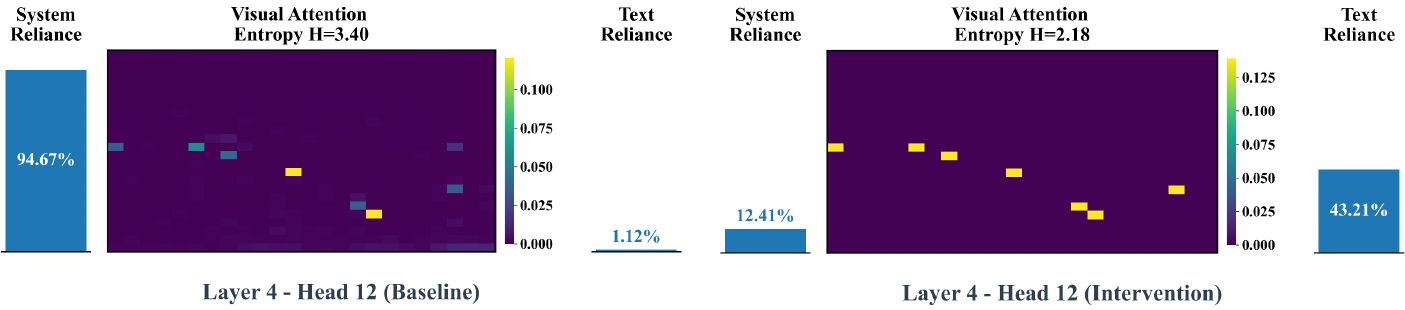}
}\\[0.45em]
\subfloat[Representative risky head (example 4).]{%
  \includegraphics[width=0.97\linewidth]{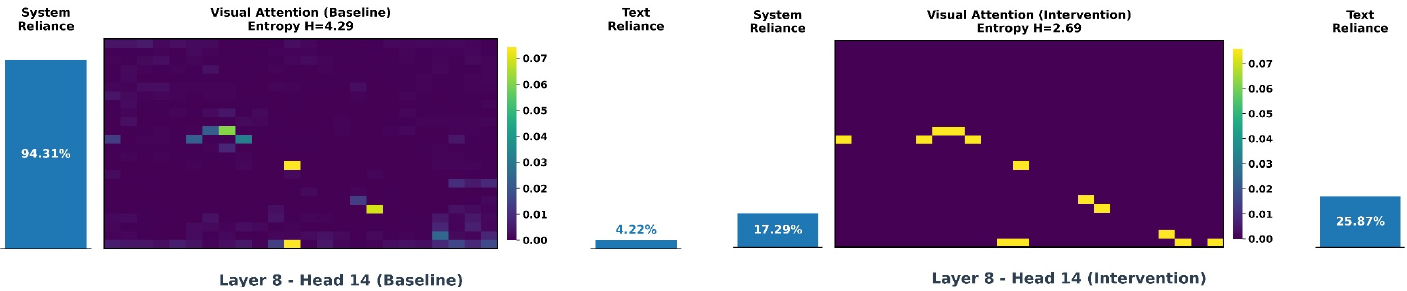}
}\\
\end{tabular}

\caption{
\textbf{Head-level structural transformation after logit-level intervention (four representative examples).}
For each risky head (selected by high joint risk $S^{(l,h)}_{\text{tail}}$), we visualize the attention snapshot at a fixed decision-critical step: \emph{system reliance} $m^{(l,h)}_{sys,\text{tail}}$ (left bar), \emph{visual attention map} over $\mathcal{I}_{vis}$ with entropy $H^{(l,h)}_{vis,\text{tail}}$ (center heatmap), and \emph{text reliance} (right bar).
Across examples, the intervention consistently suppresses pathological system/prefix lock-on and sharpens visual evidence routing, leading to a lower visual entropy and a more concentrated set of visual peaks.
These four cases illustrate that the structural reconfiguration induced by our intervention is not confined to a single head or layer, but recurs across multiple risky mediators.
}
\label{fig:app_intervention_comparison}
\vspace{-1em}
\end{figure*}

\paragraph{Visualization setup.}
Figure~\ref{fig:app_intervention_comparison} provides a multi-view diagnostic snapshot for each head at the fixed query step:
(i) System Reliance, the attention mass on $\mathcal{I}_{sys}$ (i.e., $m^{(l,h)}_{sys,\text{tail}}$);
(ii) Visual Attention Map, the distribution over $\mathcal{I}_{vis}$ together with the corresponding visual entropy (i.e., $H^{(l,h)}_{vis,\text{tail}}$); and
(iii) Text Reliance, the attention mass on the textual context.
Text reliance is shown only to illustrate redistribution after intervention and is not used in risk scoring or head selection.

\paragraph{Mechanistic interpretation.}
Our intervention applies a large negative bias to the attention logits $\mathbf{L}^{(l,h)}$ \emph{before} Softmax, which (under finite precision) drives unreliable connections to near-zero probability and forces the remaining mass to be re-normalized.
Across the four examples, we observe a consistent reconfiguration pattern:
(i) strong system/prefix lock-on is reduced;
(ii) the visual map becomes less diffused and more peak-focused, reflected by a decrease in $H^{(l,h)}_{vis,\text{tail}}$; and
(iii) the removed mass is redistributed to more plausible visual or textual anchors.
Together, these qualitative cases provide direct evidence that heads with high $S^{(l,h)}_{\text{tail}}=m^{(l,h)}_{sys,\text{tail}}\cdot H^{(l,h)}_{vis,\text{tail}}$ indeed exhibit simultaneous system hijacking and elevated visual uncertainty, and that our intervention reconstructs their routing structure toward more grounded evidence.

\section{Method}
\label{sec_app:method}

This subsection specifies how \M~\emph{executes} the causal intervention at inference time.
After identifying the risky mediators $H_R$ in our diagnosis stage, we instantiate two coupled next-token distributions at each decoding step $t$:
(i) an \emph{observational} branch $P_{obs}(\cdot)$ that follows the original computation graph, and
(ii) an \emph{interventional} branch $P_{do}(\cdot)$ that enforces $\mathbf{do}(H_R)$ by editing attention logits before Softmax on the selected heads.
Causally, the intervention aims to attenuate the shortcut pathway $\mathbf{X}_{sys}\!\rightarrow\!H_R\!\rightarrow\!Y_t$ so that the next-token decision relies less on system/prefix priors and becomes more grounded in the multimodal evidence, while the observational branch preserves the model's native linguistic manifold.
Our objective is to leverage the improved faithfulness of $P_{do}$ without over-committing to it when the intervention becomes overly restrictive.

\paragraph{Step-wise procedure.}
Concretely, at each generation step, we perform the following operations:

\begin{itemize}[leftmargin=*,itemsep=2pt,topsep=2pt]
    \item \textbf{(1) Dual forward passes (observational vs.\ interventional).}
    We compute the next-token logits from the original run to obtain $P_{obs}$.
    In parallel, we run the model again with $\mathbf{do}(H_R)$ applied on the diagnosed heads to obtain $P_{do}$.
    In practice, $\mathbf{do}(H_R)$ is implemented exactly as in Algorithm~1: on the designated intervention layers (early-to-mid range), we select Top-$\lceil k\!\cdot\!H\rceil$ heads per layer using the joint risk score and apply a negative bias to their \emph{pre-Softmax} attention logits on decision-critical queries, driving unreliable links to near-zero probability after re-normalization. Concretely, for $(l,h)\in H_R$ and $q\in\mathcal{Q}$, we use numerical logit saturation:
$\tilde{\mathbf{L}}^{(l,h)}=\Pi_{\text{dtype}}(\mathbf{L}^{(l,h)}-\gamma)$, and then $\tilde{\mathbf{A}}^{(l,h)}=\mathrm{Softmax}(\tilde{\mathbf{L}}^{(l,h)})$.

    \item \textbf{(2) Candidate truncation for stable conflict measurement.}
    Measuring divergence over the full vocabulary is dominated by the long tail of near-zero probabilities.
    We therefore construct a decision-relevant candidate set $\mathcal{V}_t$ induced by $P_{obs}$, retaining only tokens within a fixed ratio of the top-1 probability.
    This truncation makes the conflict estimate focus on the local decision boundary rather than numerical tail noise.

    \item \textbf{(3) Conflict estimation.}
    We quantify the disagreement between the observational and interventional branches using Jensen--Shannon divergence on the truncated candidates:
    \begin{equation}
   d_t=\mathrm{JSD}\!\left(P_{obs}(\cdot \mid \mathcal{V}_t)\,\|\,P_{do}(\cdot \mid \mathcal{V}_t)\right).
    \end{equation}
    A small $d_t$ indicates that the intervention stays close to the observational manifold at step $t$,
    whereas a large $d_t$ signals a strong causal perturbation that substantially reshapes the next-token preference.

    \item \textbf{(4) Conflict-gated injection.}
    We convert $d_t$ into a step-wise injection weight $\lambda_t$.
    When the two branches are consistent (low conflict), we apply a fixed gain $\alpha$ to strengthen the interventional correction.
    When they diverge (high conflict), we fall back to a softer, conflict-proportional injection, preventing the interventional branch from overwhelming the observational manifold.
\end{itemize}

\paragraph{Logit-level combination.}
Finally, we couple the two branches at the logit level and select the next token:
\begin{equation}
\mathbf{z}_{final,t}=\mathbf{z}_{obs,t}+\lambda_t \cdot \mathbf{z}_{do,t},
\qquad
y_t \sim \mathrm{Softmax}(\mathbf{z}_{final,t}),
\end{equation}
where $\lambda_t$ is determined by the conflict-gating rule described above.

\begin{center}
\small
\setlength{\tabcolsep}{6pt}
\renewcommand{\arraystretch}{1.15}
\begin{tabular}{p{0.96\linewidth}}
\hline
\textbf{Algorithm 1: \M~framework} \\ \hline
\textbf{Input:} $\mathbf{X}$, history $y_{<t}$, model $\mathcal{F}_\theta$; intervention layers $\mathcal{L}_{int}$;
decision queries $\mathcal{Q}$; per-layer ratio $k$; intervention bias $\gamma$;
conflict threshold $\tau_{\mathrm{JS}}$; gain $\alpha$; truncation ratio $\beta$. \\
\textbf{Output:} next token $y_t$ (repeat for $t=1,\dots,T$). \\
\textbf{(1) Pass-1.} Compute logits $\mathbf{z}^{obs}_t=\mathcal{F}_\theta(\mathbf{X},y_{<t})$ and
$P_{obs}=\mathrm{Softmax}(\mathbf{z}^{obs}_t)$. \\
\textbf{(2) Pass-2 (with intervention on $\mathcal{L}_{int}$).}
Run $\mathcal{F}_\theta(\mathbf{X},y_{<t})$ again; for each layer $l\in\mathcal{L}_{int}$: \\
\quad (a) On decision queries $\mathcal{Q}$, compute per-head metrics
$m^{(l,h)}_{sys}$ and $H^{(l,h)}_{vis}$, and score
$S^{(l,h)}=m^{(l,h)}_{sys}\cdot H^{(l,h)}_{vis}$. \\
\quad (b) Select risky heads $\mathcal{H}_l=\text{Top-}\lceil k\cdot N_h\rceil$ by $S^{(l,h)}$. \\
\quad (c) \textbf{Pre-Softmax logit intervention:} for $h\in\mathcal{H}_l$ and $q\in\mathcal{Q}$,
apply a negative bias $-\gamma$ to the corresponding attention logits. \\
Obtain interventional logits $\mathbf{z}^{do}_t$ and $P_{do}=\mathrm{Softmax}(\mathbf{z}^{do}_t)$. \\
\textbf{(3) Conflict gating.}
Construct
$\mathcal{V}_t=\{y\in\mathcal{V}\mid P_{obs}(y)\ge \beta\cdot \max_w P_{obs}(w)\}$ and compute
\[
d_t=\mathrm{JSD}\!\left(P_{obs}(\cdot\mid \mathcal{V}_t)\,\|\,P_{do}(\cdot\mid \mathcal{V}_t)\right).
\]
Set $\lambda_t=\alpha$ if $d_t<\tau_{\mathrm{JS}}$, otherwise $\lambda_t=d_t$. \\
\textbf{(4) Logit fusion and selection.}
Fuse
$\mathbf{z}_{final,t}=\mathbf{z}_{obs,t}+\lambda_t\cdot \mathbf{z}_{do,t}$ and select
$y_t \sim \mathrm{Softmax}(\mathbf{z}_{final,t})$. \\ \hline
\end{tabular}
\end{center}

Algorithm~1 summarizes the step-wise inference procedure.

\section{Detailed Configurations and Experimental Results}

\subsection{Models and Baselines}
\label{app:baselines}
We select LVLMs that represent diverse architectural paradigms to ensure the generalizability of our method:
\begin{itemize}
    \item \textbf{LLaVA-1.5}~\cite{liu2023visual}: A widely-used general-purpose baseline that connects a CLIP-ViT-L/14 encoder with the Vicuna LLM via a two-layer MLP projection.
    \item \textbf{Shikra}~\cite{chen2023shikra}: A structured LVLM specialized for referential dialogue and fine-grained object grounding, handling bounding box inputs/outputs.
    \item \textbf{InstructBLIP}~\cite{dai2023instructblip}: An instruction-tuned model utilizing a Q-Former to compress visual features into soft queries for the LLM.
\end{itemize}

For baselines, we compare against the following inference-time interventions:
\begin{itemize}
    \item \textbf{ICD}~\cite{ICD}: Constructs contrastive instruction branches to estimate and suppress language priors.
    \item \textbf{VCD}~\cite{VCD}: Introduces visual noise to amplify hallucination-prone logits via contrastive decoding.
    \item \textbf{OPERA}~\cite{huang2024opera}: A beam-search-based method that detects ``over-trust'' attention patterns and applies a rollback penalty.
    \item \textbf{SID}~\cite{SID}: Reweights candidate tokens dynamically using self-contrastive signals to prevent error amplification.
    \item \textbf{CausalMM}~\cite{CausalMM}: Applies counterfactual reasoning on both encoder and decoder sides to disentangle spurious correlations.
\end{itemize}

\subsection{Hyperparameters and Hardware}
\label{app:implementation}
\textbf{Sampling Strategy.} To simulate realistic generation scenarios, we use Nucleus Sampling with top-$p=0.9$, temperature $T=1.0$, and a maximum length of 512 tokens. No repetition or length penalties are applied. Note that OPERA is evaluated using its official beam-search configuration (num\_beams=5) as it is incompatible with standard sampling.

\textbf{Method-Specific Parameters.}
Our method involves four key hyperparameters: the per-layer head suppression ratio $k$, the conflict threshold $\tau_{\mathrm{JS}}$, the consensus amplification factor $\alpha$, and the truncation ratio $\beta$ used in conflict estimation.

We fix $\alpha = 2$ across all backbones. The remaining hyperparameters $(k, \tau_{\mathrm{JS}}, \beta)$ are selected in a model-specific manner via grid search on a held-out validation set. Notably, the search consistently selects the same $\beta$ across all evaluated backbones, and we therefore use $\beta = 0.1$ for LLaVA-1.5, InstructBLIP, and Shikra.

The optimal configurations are:
\begin{itemize}
    \item \textbf{LLaVA-1.5:} $k = 0.45$, $\tau_{\mathrm{JS}} = 0.2$, $\beta = 0.1$.
    \item \textbf{InstructBLIP:} $k = 0.4$, $\tau_{\mathrm{JS}} = 0.2$, $\beta = 0.1$.
    \item \textbf{Shikra:} $k = 0.4$, $\tau_{\mathrm{JS}} = 0.2$, $\beta = 0.1$.
\end{itemize}

We report the best results averaged over $\text{10}$ independent runs. Statistical significance is determined by a two-sided $t$-test ($p < 0.05$). All experiments are conducted on $8\times$ NVIDIA A100 (40GB) GPUs using PyTorch and HuggingFace Transformers.

\subsection{Detailed Metrics and Protocols}

\paragraph{(1) POPE.}
This metric evaluates object existence through a series of binary (Yes/No) questions (e.g., ``\textit{Is there a [object] in the image?}''), thereby measuring the model's propensity to fabricate non-existent visual evidence. POPE consists of three distinct sampling configurations to assess different facets of model reliability:

\begin{itemize}
    \item \textbf{Random:} Targets are sampled randomly with broad category coverage to examine the model's general recognition and grounding capabilities across diverse objects.
    \item \textbf{Popular:} Categories with higher frequencies in the training distribution are selected to observe if the model is more stable and less prone to hallucination when dealing with ``common and familiar'' objects.
    \item \textbf{Adversarial:} Categories that are frequently misreported or confused by LVLMs are selected, increasing the evaluation difficulty and revealing the model's robustness against hallucination under ambiguous or interfering inputs.
\end{itemize}

By focusing on binary probing, this benchmark bypasses the complexities associated with parsing open-ended generated captions, ensuring a stable, fair, and adaptable evaluation process. We report both Accuracy and the F1-score to quantify performance. The F1-score serves as a balanced harmonic mean between Precision and Recall, defined as:
\begin{equation}
    \text{Recall} = \frac{\text{Correctly identified objects}}{\text{Ground-truth objects}},
\end{equation}
\begin{equation}
    \text{Precision} = \frac{\text{Correctly identified objects}}{\text{Total generated objects}},
\end{equation}
\begin{equation}
    \text{F1} = 2 \times \frac{\text{Precision} \times \text{Recall}}{\text{Precision} + \text{Recall}}.
\end{equation}

In our experimental framework, Recall characterizes the proportion of ground-truth objects successfully retrieved by the model from the visual evidence $\mathbf{X}_{vis}$, while Precision measures the ratio of generated objects that actually exist in the image rather than being hallucinations. As the harmonic mean of both, the F1-score provides a robust holistic measure of generation quality. Consequently, Accuracy and F1-score constitute our standard baseline framework for assessing the model's overall efficacy in multimodal grounding.

\paragraph{(2) CHAIR.}
The \textbf{CHAIR}~\cite{chair} benchmark consists of two primary metrics, $\text{CHAIR}_I$ and $\text{CHAIR}_S$, which measure object hallucinations in image captioning at the instance and sentence levels, respectively. Specifically, the instance-level metric $\text{CHAIR}_I$ calculates the proportion of hallucinated objects relative to all mentioned objects in the generated captions. The sentence-level metric $\text{CHAIR}_S$ reflects the proportion of generated sentences that contain at least one hallucination. 

To ensure that our intervention $\mathbf{do}(H_R)$ does not merely reduce hallucinations by excessively suppressing the generation of fine-grained details, we further incorporate Recall and F1-score as indicators of semantic completeness. These metrics verify that the performance gains are not achieved through an ``evasion strategy'' but through reliable visual grounding. The metrics follow the formulations below:
\begin{equation}
\text{CHAIR}_I = \frac{|\{\text{hallucinated objects}\}|}{|\{\text{all mentioned objects}\}|},
\end{equation}
\begin{equation}
\text{CHAIR}_S = \frac{|\{\text{sentences with hallucinations}\}|}{|\{\text{total sentences}\}|}.
\end{equation}
The robustness of the captions is measured via:
\begin{equation}
\text{Recall} = \frac{|\{\text{accurately mentioned objects}\}|}{|\{\text{ground-truth objects}\}|},
\end{equation}
\begin{equation}
\text{Precision} = \frac{|\{\text{all mentioned objects}\} \cap \{\text{ground-truth objects}\}|}{|\{\text{all mentioned objects}\}|},
\end{equation}
\begin{equation}
\text{F1} = 2 \times \frac{\text{Precision} \times \text{Recall}}{\text{Precision} + \text{Recall}}.
\end{equation}
Collectively, $\text{CHAIR}_I$, $\text{CHAIR}_S$, and the F1-score constitute our comprehensive evaluation framework for the captioning task.

\paragraph{(3) MME.}
The \textbf{MME}~\cite{mme} benchmark pairs each image with two semantically similar questions whose ground-truth answers are ``Yes'' and ``No,'' respectively. Evaluation is conducted using two metrics: Accuracy and Accuracy+. 

\begin{itemize}
    \item \textbf{Accuracy} is calculated at the question granularity: a correct response to any single question contributes to the score. 
    \item \textbf{Accuracy+} is calculated at the image granularity: a sample is counted as correct only if the model correctly answers \emph{both} the ``Yes'' and ``No'' questions associated with the same image. 
\end{itemize}
This stricter requirement ensures that the model truly perceives the visual evidence $\mathbf{X}_{vis}$ rather than relying on language priors. The final MME Score is defined as the sum of these metrics:
\begin{equation}
\text{Accuracy} = \frac{\sum_{i \in \mathcal{I}} \mathbf{1}[f(i, q_{yes}) = \text{``Yes''}] + \sum_{i \in \mathcal{I}} \mathbf{1}[f(i, q_{no}) = \text{``No''}]}{2|\mathcal{I}|},
\end{equation}
\begin{equation}
\text{Accuracy}^+ = \frac{\sum_{i \in \mathcal{I}} \mathbf{1}[f(i, q_{yes}) = \text{``Yes''} \land f(i, q_{no}) = \text{``No''}]}{|\mathcal{I}|},
\end{equation}
\begin{equation}
\text{MME Score} = \text{Accuracy} + \text{Accuracy}^+.
\end{equation}

\paragraph{(4) GPT-4V Assisted Evaluation.}
\label{app:gpt4v}
While CHAIR and POPE effectively identify object hallucinations, they provide limited insight into the overall linguistic quality and descriptive richness of open-ended outputs. To complement these automatic metrics, we conduct a GPT-4V--assisted evaluation by following an established protocol from prior work~\cite{huang2024opera,yang2023dawnlmmspreliminaryexplorations}. Specifically, GPT-4V acts as a multimodal judge and assigns scores under two criteria: Accuracy (factual consistency with respect to objects, attributes, and spatial/relational correctness) and Detailedness (the richness and precision of \emph{correctly grounded} visual details). Leveraging its advanced perception, GPT-4V can capture subtle errors in color, spatial positioning, and logical relationships between objects.
We conduct this evaluation on a curated subset of 50 images from the MS-COCO 2014 validation set, used solely as complementary evidence rather than a primary benchmark. For each image--model pair, we generate descriptions with a standardized prompt and evaluate both the original backbone output and the output produced with our method. GPT-4V inference is performed with \texttt{max\_tokens}=512 and \texttt{temperature}=0.2. The full prompt and required output format are provided in Table~\ref{tab:gpt4v_prompt}.

\begin{table*}[t]  % 使用table*实现跨栏
\centering  
\begin{tabular}{p{0.95\textwidth}}  % 使用textwidth适应跨栏宽度
\hline  
\textbf{GPT-4V Prompt} \\    
\hline    
You are required to score the performance of two AI assistants in describing a given image. You should pay extra attention to the hallucination, which refers to the part of descriptions that are inconsistent with the image content, such as claiming the existence of something not present in the image or describing incorrectly in terms of the counts, positions, or colors of objects in the image. Please rate the responses of the assistants on a scale of $1$ to $10$, where a higher score indicates better performance, according to the following criteria: \par  

\textbf{1. Accuracy:} Evaluate whether the response is accurate and faithful to the actual image content. Focus on identifying any hallucinations including non-existent objects, incorrect attributes (colors, sizes, materials), wrong quantities, false spatial relationships, or activities that are not happening. Responses with fewer hallucinations and higher fidelity to the image should receive higher scores. \par  

\textbf{2. Detailedness:} Assess whether the response provides rich and informative details about the image. Consider the completeness of the description, coverage of important visual elements, and the depth of observations. Note that hallucinated content does NOT count as valid details -- only accurate information contributes to this score. \par  

Please output the scores for each criterion, containing only two values indicating the scores for Assistant 1 and 2, respectively. The two scores are separated by a space. Following the scores, please provide an explanation of your evaluation, avoiding any potential bias and ensuring that the order in which the responses were presented does not affect your judgment. \par  

[Assistant 1] \par  
\{Response of Assistant 1\} \par  
[End of Assistant 1] \par  

[Assistant 2] \par  
\{Response of Assistant 2\} \par  
[End of Assistant 2] \par  

\textbf{Output format:} \par  
Accuracy: \textless score\_1\textgreater~\textless score\_2\textgreater \par  
Reason: \textless your explanation\textgreater \par  
Detailedness: \textless score\_1\textgreater~\textless score\_2\textgreater \par  
Reason: \textless your explanation\textgreater \par   \\
\hline    
\end{tabular}  
\caption{The prompt used for GPT-4V evaluation.}  
\label{tab:gpt4v_prompt}  
\end{table*}

\subsection{Additional Ablation Studies}\label{app:add_ablation}

\paragraph{Efficiency and Latency Analysis.}\ 
Table~\ref{tab:efficiency} compares the inference cost on LLaVA-1.5 alongside POPE Adversarial performance. \M~achieves a superior Pareto trade-off, maintaining the same latency regime as VCD/SID ($\approx$$200$~ms/token) while reaching the highest accuracy ($81.93$\%). For 10-token generation, \M~incurs only modest overhead ($1,040$~ms), whereas search-based methods like OPERA exhibit significantly higher latency ($2,560$~ms) due to beam search and iterative rollback.
These results confirm that our gains stem from a logit-level intervention within the same dual-pass contrastive decoding regime as VCD/SID, rather than increased decoding depth or search budgets. In practice, the system-prompt and visual-token KV cache can be shared between branches, keeping the overhead in the same latency class while providing a stronger balance between faithfulness and efficiency.

\begin{table}[ht]
\centering
% \scriptsize  % 比\footnotesize更紧凑，核心数据仍清晰
\setlength{\tabcolsep}{5pt}  % 列间距从6pt压至3pt，大幅缩宽
\renewcommand{\arraystretch}{1}  % 行高收紧，减少纵向留白
\begin{tabular}{c|c|c|c}
\hline
\textbf{Method} & 1/token (ms) & 10/token (ms) & ACC (\%) \\
\hline
Sampling & 110 & 437 & 76.77 \\
VCD      & 204 & 893 & 75.33 \\
OPERA    & -   & 2560& 81.13 \\
SID      & 202 & 890 & 81.10 \\
\hline
\rowcolor{gray!20} \textbf{Ours} & 206 & 1040& \textbf{81.93} \\
\hline
\end{tabular}
\caption{Efficiency comparison. Inference latency (ms) for 1-token and 10-token generation. OPERA is not evaluated in the 1-token setting (marked as -).}
% \vspace{-3mm}
\label{tab:efficiency}
\end{table}

\subsubsection{\texorpdfstring{Ablation Study on $\alpha$}{Ablation Study on alpha}}
\label{app:ablation_alpha}

% \begin{minipage}{0.8\linewidth}
% \centering
% \small
% \begin{tabular}{c|cc}
% \hline
% $\alpha$ & Mean Acc $\uparrow$ & Mean F1 $\uparrow$ \\
% \hline
% 0 & 81.75 & 82.90 \\
% 1 & 85.33 & 85.35 \\
% 2 & 85.97 & \textbf{85.77} \\
% 3 & \textbf{85.98} & 85.53 \\
% \hline
% \end{tabular}

% \vspace{2pt}
% \captionof{table}{Sensitivity of the enhancement factor on POPE.}
% \label{tab:ablation_alpha}
% \end{minipage}

\begin{table}[h]
\centering
\setlength{\tabcolsep}{4pt}
\begin{tabular}{c|cc}
\toprule
$\alpha$ & Mean Acc$\uparrow$ & Mean F1$\uparrow$ \\
\midrule
0 & 81.75 & 82.90 \\
1 & 85.33 & 85.35 \\
2 & 85.97 & \textbf{85.77} \\
3 & \textbf{85.98} & 85.53 \\
\bottomrule
\end{tabular}
\caption{Sensitivity of the enhancement factor $\alpha$ on POPE.}
\label{tab:ablation_alpha}
\end{table}

As shown in Table~\ref{tab:ablation_alpha}, we conduct a sensitivity analysis of the enhancement factor $\alpha$ on POPE to determine a default setting that can be reused across backbones. Concretely, we vary $\alpha$ on LLaVA-1.5 while keeping all other hyperparameters fixed, and report the averaged \textit{Accuracy} and \textit{F1} over the Random/Popular/Adversarial splits. We observe that performance changes smoothly within a reasonably wide range of $\alpha$, with $\alpha=2$ achieving the best or near-best averaged performance. When further increasing $\alpha$, the gains exhibit diminishing returns and slightly regress on some splits, suggesting that overly strong consensus enhancement may bias the binary verification toward a more \emph{robust but conservative} decision behavior, which is detrimental to overall F1. Based on this trend, we set $\alpha=2$ as the default and keep it fixed for all subsequent backbones (including InstructBLIP and Shikra), reducing model-specific tuning freedom and verifying the transferability and robustness of this choice.

\subsubsection{Ablation Study on InstructBLIP}
\begin{figure}[htbp]
    \centering
    \includegraphics[width=0.8\linewidth]{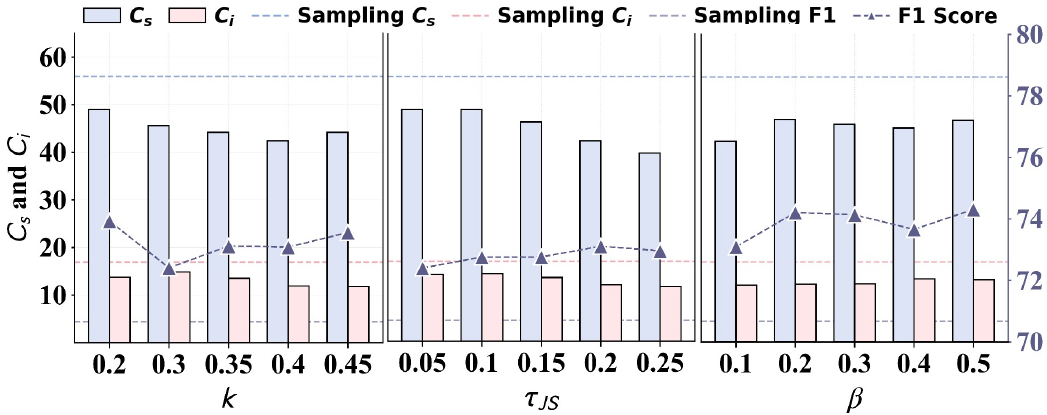} 
    \caption{Parameter sensitivity analysis on InstructBLIP. Impact of $k$, $\tau_{\mathrm{JS}}$, and $\beta$ on captioning performance (CHAIR$_S$, CHAIR$_I$, and F1), evaluated on 500 COCO samples.}
    \label{fig:parameter_instructblip}
\end{figure}

We analyze the sensitivity of key hyperparameters on InstructBLIP. Since $\alpha$ is fixed globally to $\alpha=2$ across all backbones (see Appendix~\ref{app:ablation_alpha}), we focus on the three parameters that govern intervention strength and conflict gating: the per-layer head suppression ratio $k$, the JSD conflict threshold $\tau_{\mathrm{JS}}$, and the truncation ratio $\beta$. Overall, these parameters mainly control the trade-off between \emph{reliability} (lower hallucination, reflected by CHAIR$_S$/CHAIR$_I$) and \emph{informativeness} (semantic coverage, reflected by F1), with $\tau_{\mathrm{JS}}$ serving as the primary knob.

\paragraph{Impact of $\tau_{\mathrm{JS}}$.}
As $\tau_{\mathrm{JS}}$ increases, the decoding procedure more frequently follows the intervention-dominant path at uncertain steps, which consistently reduces hallucination errors (lower CHAIR$_S$ and CHAIR$_I$). However, overly large $\tau_{\mathrm{JS}}$ yields diminishing returns and may slightly regress F1, indicating a conservative bias that prioritizes safety over semantic coverage. This suggests that $\tau_{\mathrm{JS}}$ should be set within a moderate range to suppress uncertainty-driven deviations without over-stabilizing the output.

\paragraph{Impact of $k$.}
The ratio $k$ controls the sparsity budget of head-level intervention per layer. Varying $k$ results in relatively smooth changes in both CHAIR and F1: moderate $k$ values typically improve CHAIR$_S$/CHAIR$_I$ without harming F1, while overly aggressive intervention (large $k$) brings limited additional gains and can slightly reduce F1, suggesting over-contraction of the effective generation space.

\paragraph{Impact of $\beta$.}
The parameter $\beta$ controls the truncation strength in conflict estimation. InstructBLIP is relatively robust to $\beta$ within a broad range, where performance varies smoothly and improvements on CHAIR metrics remain stable. Extreme truncation may narrow the effective candidate space and lead to diminishing returns, occasionally accompanied by a slight F1 decrease.

\paragraph{Selection of Intervention Layers in InstructBLIP.}
\label{app:layer_selection_instructblip}

For InstructBLIP, we set the intervention range to layers 4--10, adhering to our core finding that intervention must occur in the early-to-mid stages to suppress \emph{risky mediators} before erroneous evidence chains become consolidated in subsequent generation.

The distinction lies in InstructBLIP's architecture: cross-modal information is first compressed into a compact set of visual representations via the Q-Former's learnable queries before fusion with the LLM. Consequently, the ``effective visual evidence'' enters the language decoder in a manner that favors early information aggregation followed by progressive verbalization. 
\begin{itemize}
    \item \textbf{Early Layers ($<4$):} Intervening too early often acts on the stage before multimodal fusion is fully realized, resulting in unstable gains.
    \item \textbf{Latter Layers ($>10$):} Intervening in later stages primarily affects linguistic expression and decoding convergence, offering limited help in correcting biases formed during the initial fusion stage, and potentially leading to overly conservative outputs.
\end{itemize}
By targeting the 4--10 layer range, we effectively balance the suppression of uncertainty-driven shortcuts with the preservation of the model's inherent linguistic fluency.

\subsubsection{Ablation Studies on Shikra}
\label{app:ablation_shikra}
\begin{figure}[htbp]
    \centering
    \includegraphics[width=0.8\linewidth]{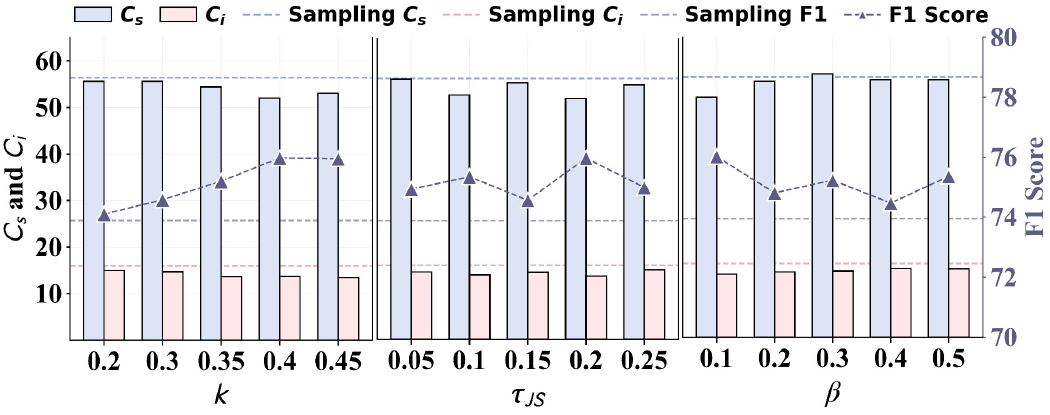} 
    \caption{Parameter sensitivity analysis on Shikra. Impact of $k$, $\tau_{\mathrm{JS}}$, and $\beta$ on captioning performance (CHAIR$_S$, CHAIR$_I$, and F1), evaluated on 500 COCO samples.}
    \label{fig:parameter_shikra}
\end{figure}

As shown in Figure~\ref{fig:parameter_shikra}, we examine the sensitivity of key hyperparameters on Shikra. Since $\alpha$ is fixed globally to $\alpha=2$ across all backbones (see Appendix~\ref{app:ablation_alpha}), we focus on the three parameters that directly govern intervention strength and conflict gating: the per-layer head suppression ratio $k$, the JSD conflict threshold $\tau_{\mathrm{JS}}$, and the truncation ratio $\beta$. Overall, these parameters mainly control the trade-off between \emph{reliability} (lower hallucination, reflected by CHAIR$_S$/CHAIR$_I$) and \emph{informativeness} (semantic coverage, reflected by F1), with $\tau_{\mathrm{JS}}$ remaining the most influential knob.

\paragraph{Impact of $\tau_{\mathrm{JS}}$.}
As $\tau_{\mathrm{JS}}$ increases, the decoding procedure more often follows the intervention-dominant path at uncertain steps, yielding consistent reductions in hallucination metrics (lower CHAIR$_S$ and CHAIR$_I$). However, overly large $\tau_{\mathrm{JS}}$ introduces diminishing returns and may slightly reduce F1, indicating a conservative bias that favors safer generation over semantic coverage. This suggests that $\tau_{\mathrm{JS}}$ should be set in a moderate range to suppress uncertainty-driven deviations without over-stabilizing the output.

\paragraph{Impact of $k$.}
The ratio $k$ controls the sparsity budget of head-level intervention per layer. Figure~\ref{fig:parameter_shikra} shows that varying $k$ leads to relatively smooth changes in both CHAIR and F1. Moderate $k$ values typically provide a favorable balance, improving CHAIR$_S$/CHAIR$_I$ without harming F1, while overly aggressive intervention (large $k$) offers limited additional gains and can slightly regress F1.

\paragraph{Impact of $\beta$.}
The parameter $\beta$ controls the truncation strength in conflict estimation. Performance varies smoothly across a broad range of $\beta$, suggesting that Shikra is relatively robust to this parameter. Moderate $\beta$ values achieve stable improvements on CHAIR metrics while maintaining F1, whereas extreme truncation may narrow the effective candidate space and yield diminishing returns.

\paragraph{Selection of Intervention Layers in Shikra.}
\label{app:layer_selection_shikra}

For Shikra, we set the intervention range to layers 3--10. The core rationale remains consistent with our previous findings: the erroneous evidence chains of hallucinations are typically established and progressively amplified during the early-to-mid stages of decoding. Intervening only in the later stages is usually insufficient for timely correction. Therefore, we place the intervention within the ``early-to-mid'' window to suppress the influence of risky mediators on subsequent generation as early as possible.

Compared to LLaVA-1.5 (e.g., layers 2--15), Shikra's optimal window is slightly shifted forward and is narrower, primarily due to differences in model architecture and task format.
\begin{itemize}
    \item \textbf{Early Structural Routing:} As a model designed for referential grounding, Shikra's input sequences contain more explicit region/object-related tokens. Cross-modal alignment forms strong structural routing in earlier layers.
    \item \textbf{Generative Convergence:} Subsequent layers tend toward the ``convergence'' of linguistic generation and instruction execution based on the established alignment. Intervening at this stage is more likely to disrupt stable generation while providing limited help in correcting errors formed during early alignment.
\end{itemize}

Based on this divergence, while maintaining the principle of ``early-to-mid stage intervention,'' we set Shikra’s intervention range to layers 3--10 to better fit its internal dynamics where alignment is established early and generation converges later.

\begin{table}[htbp]
    \centering
    \setlength{\tabcolsep}{4pt}
    \resizebox{0.7\linewidth}{!}{
    \begin{tabular}{c|ccc|ccc|ccc}
        \toprule
        \multirow{2}*{\textbf{Method}} & 
        \multicolumn{3}{c|}{\textbf{\emph{LLAVA-1.5}}} & 
        \multicolumn{3}{c|}{\textbf{\emph{InstructBLIP}}} &
        \multicolumn{3}{c}{\textbf{\emph{Shikra}}} \\
        ~ & Ran$\uparrow$ & Pop$\uparrow$ & Adv$\uparrow$ & Ran$\uparrow$ & Pop$\uparrow$ & Adv$\uparrow$ & Ran$\uparrow$ & Pop$\uparrow$ & Adv$\uparrow$ \\
        \midrule
        Sampling & 85.59 & 83.40 & 79.06 & 86.14 & 81.55 & 78.80 & 82.17 & 81.06 & 77.44 \\
        VCD & 86.83 & 82.05 & 78.00 & 85.70 & 81.12 & 79.87 & 79.13 & 81.12 & 75.90 \\
        ICD & 86.46 & 84.18 & 79.83 & 87.90 & 81.74 & 79.82 & 80.14 & 80.20 & 77.89 \\
        OPERA & 88.72 & \textbf{86.59} & 81.87 & 89.43 & 83.19 & \textbf{82.34} & 83.66 & \textbf{83.21} & \textbf{80.01} \\
        CausalMM & 88.63 & 86.17 & 81.93 & 87.83 & 83.27 & 82.23 & 82.53 & 82.89 & 79.47 \\
        SID & 88.18 & 85.21 & 81.69 & 86.10 & 83.51 & 80.34 & 82.28 & 79.97 & 80.20 \\
        \hline
        \rowcolor{gray!20}\textbf{\M~(Ours)} & \textbf{88.74} & 86.30& \textbf{82.26} & \textbf{89.44} & \textbf{83.77} & 81.56 & \textbf{84.68} & 82.27 & 79.55 \\
        \bottomrule
    \end{tabular}
    }
    \caption{POPE F1 Score on the Random/Popular/Adversarial splits for three LVLM backbones (LLaVA-1.5, InstructBLIP, and Shikra). Higher is better.}

    \label{tab_app:pope}
\end{table}

\subsection{POPE F1 Results}
In the main paper, we report Accuracy on POPE as the primary metric to emphasize hallucination suppression.
To provide a complementary view of the precision--recall trade-off, we additionally report F1 scores on all POPE splits in Table~\ref{tab_app:pope}.
Overall, our method remains \emph{competitive} in F1 across backbones and splits: it achieves the best F1 on LLaVA-1.5 (Random/Adversarial) and on InstructBLIP (Random/Popular), while maintaining comparable performance on the remaining settings.
These results suggest that the accuracy gains reported in the main paper are not obtained by a degenerate conservative strategy that trivially avoids positive answers, but are accompanied by a balanced precision--recall behavior.

\subsection{Additional MME Results}
\label{app:mme}

\begin{figure}[t]
\centering
\includegraphics[width=0.48\textwidth]{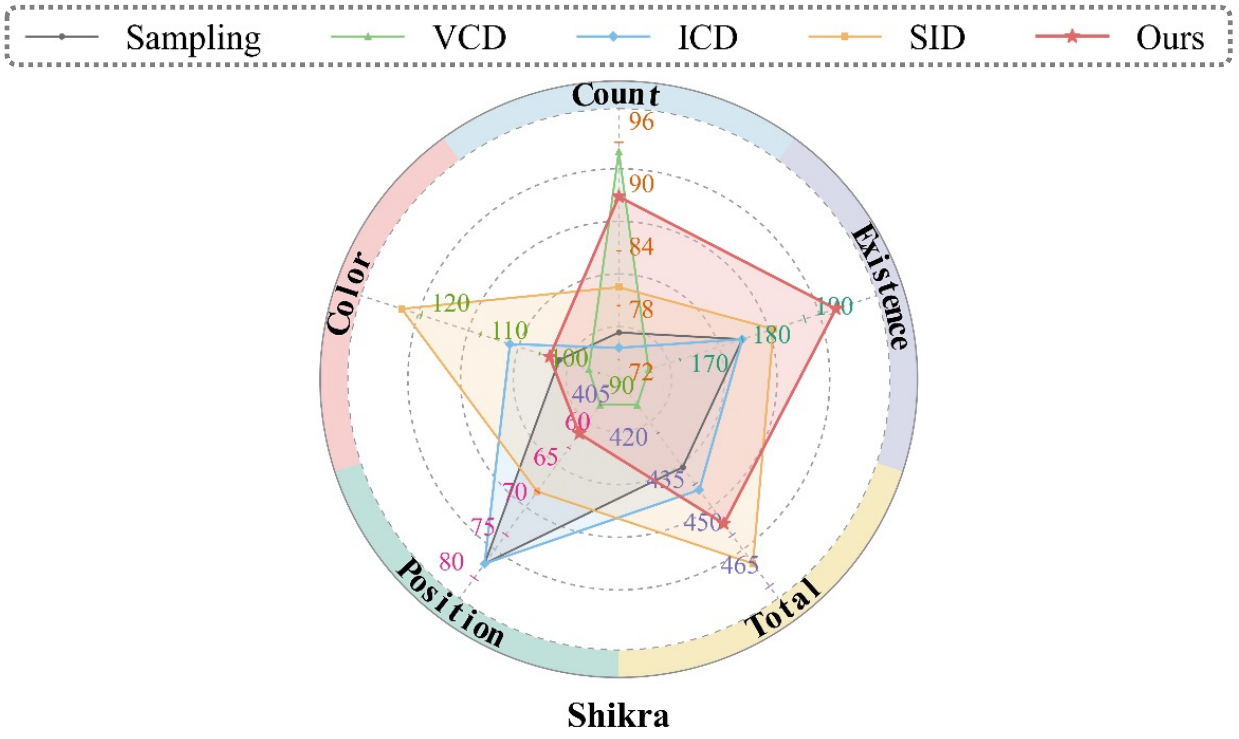}
\caption{
Performance on the MME benchmark. Higher scores indicate better effectiveness.}
\vspace{-0.5em}
\label{fig_app:MME}
\end{figure}

As shown in Figure~\ref{fig_app:MME}, in \textsc{Shikra}, \M\ still achieves an overall improvement on the MME benchmark, increasing the total score from $430.00$ to $446.63$, which validates the effectiveness of our structural intervention under this backbone.
Specifically, \M\ yields stable gains on object-level dimensions, including \textit{Existence} ($180 \to 195$) and \textit{Count} ($75 \to 90$), and also brings a slight improvement on \textit{Color} ($96.67 \to 98.33$).
In contrast, \textit{Position} exhibits a noticeable drop ($78.33 \to 63.33$), suggesting that Shikra is more sensitive to attention-structure changes for spatial attribute verification, and the benefits of structural intervention can vary across fine-grained dimensions.

\subsection{Results under Greedy Decoding}
\label{app:Greedy_Decoding}

To rule out potential confounding effects introduced by stochastic decoding (e.g., sampling temperature or nucleus sampling), we further evaluate all methods under greedy decoding on LLaVA-1.5. In this setting, the next token is selected deterministically by maximizing the conditional probability at each step, thereby eliminating randomness from the generation process.

\begin{table}[t]
\centering
\small
\setlength{\tabcolsep}{5.5pt}
\renewcommand{\arraystretch}{1.05}
% 1. 将第一个 l 改为 c (让第一列内容居中)
\begin{tabular}{c|cc|cc|cc|cc}
\hline
% 2. 使用 \multirow{2}* 让 Method 垂直居中跨越两行
\multirow{2}*{\textbf{Method}} & \multicolumn{2}{c|}{\textbf{Random}} & \multicolumn{2}{c|}{\textbf{Popular}} & \multicolumn{2}{c|}{\textbf{Adversarial}} & \multicolumn{2}{c}{\textbf{CHAIR}} \\
% 第二行第一列留空 (~)
~ & Acc $\uparrow$ & F1 $\uparrow$ & Acc $\uparrow$ & F1 $\uparrow$ & Acc $\uparrow$ & F1 $\uparrow$ & CHAIR$_S$ $\downarrow$ & CHAIR$_I$ $\downarrow$ \\
\hline
Greedy        & 89.33 & \textbf{89.28} & 85.93 & 86.33 & 79.10 & 80.95 & 50.9 & 15.40 \\
VCD           & 87.97 & 87.84 & 85.33 & 85.67 & 78.43 & 80.26 & 54.6 & 16.60 \\
OPERA         & 89.27 & 88.72 & 86.80 & 86.59 & 81.13 & 81.87 & 49.0 & 13.52 \\
SID           & 89.33 & 89.10 & 86.47 & 86.25 & 81.77 & 82.31 & 48.0 & 13.60 \\
\hline
\rowcolor{gray!20} \textbf{\M~(Ours)} & \textbf{89.43} & 88.81 & \textbf{87.07} & \textbf{86.60} & \textbf{82.43} & \textbf{82.64} & \textbf{42.6} & \textbf{12.16} \\
\hline
\end{tabular}
\caption{\textbf{Greedy decoding results on POPE and CHAIR.} We report split-wise POPE Accuracy and F1 on Random/Popular/Adversarial, together with CHAIR$_S$ and CHAIR$_I$. Since greedy decoding is used as a diagnostic control to eliminate sampling stochasticity, we omit averaged scores and focus on split-level robustness across difficulty regimes.}
\label{tab:greedy_pope_chair}
\end{table}

As shown in Table~\ref{tab:greedy_pope_chair}, our method remains consistently strong in the fully deterministic setting. In particular, it achieves the best performance on the challenging POPE-Adversarial split and simultaneously yields the lowest CHAIR$_S$ and CHAIR$_I$, indicating fewer hallucinated objects at both the sentence and instance levels. These improvements persist without any sampling variance, suggesting that the gains are not attributable to temperature tuning or favorable sampling randomness, but stem from the proposed structural intervention.

These results provide strong evidence that our improvements stem from \emph{structural intervention on high-risk mediators}, rather than from stochastic decoding effects.
In particular, even in the absence of sampling diversity, our method effectively suppresses hallucination while maintaining semantic coverage, as reflected by improvements on both POPE and CHAIR metrics.

\subsection{Mitigating Generation Degradation via Conflict-Gated Cooperation}
\label{app:cooperation_rationale}

\begin{figure}[ht]
    \centering
    \includegraphics[width=0.6\linewidth]{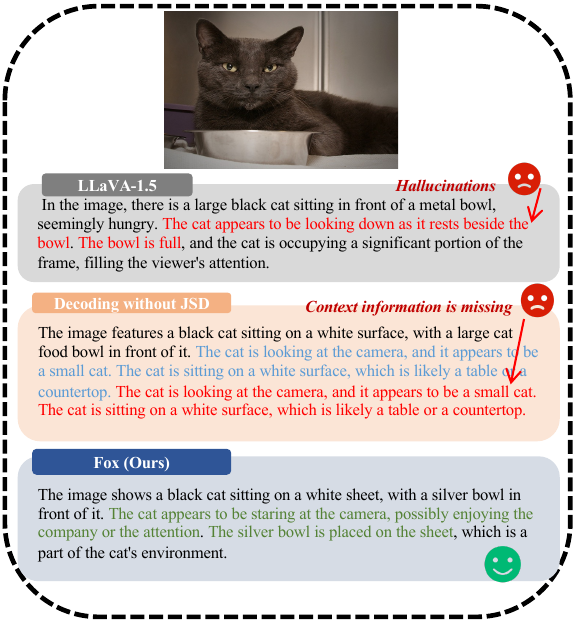} 
    \caption{
    \textbf{The role of JSD-based conflict gating.}
Without JSD gating, always applying the intervention leads to severe generation degradation due to excessive context suppression.
Conversely, an overly large JSD threshold biases the decoding toward an overly conservative regime, reducing semantic coverage.
A moderate JSD threshold enables adaptive cooperation between the interventional and observational branches, achieving a balanced trade-off between factual reliability and generation quality.
}
    \label{fig:context_loss}
    \vspace{-3mm}
\end{figure}

As demonstrated in our previous analysis, the structural intervention on risky mediators effectively severs the shortcut path $\mathbf{X}_{sys} \to H_R \to Y_t$, thereby promoting visual grounding. However, as illustrated in Figure~\ref{fig:context_loss}, this ``de-priorization'' process can lead to unintended consequences. While the interventional branch $P_{do}$ achieves high factual accuracy by strictly relying on stable visual evidence, it simultaneously compresses the model's linguistic expressive capacity, occasionally resulting in repetitive patterns or a lack of semantic richness, a common challenge for attention-based suppression methods.

\paragraph{Rationale for JSD-based Cooperative Decoding.}
To address this, we employ a Conflict-Gated Cooperative Decoding strategy (Insight III) based on Jensen-Shannon Divergence ($d_t$). The rationale for using JSD as the gating mechanism is twofold:
\begin{itemize}[leftmargin=*,itemsep=2pt,topsep=2pt,parsep=0pt]
    \item \textbf{Real-time Conflict Detection:} $d_t$ serves as a sensitive probe to detect when the interventional branch significantly deviates from the observational manifold. High divergence suggests that the model is at a critical decision point where the intervention might be over-suppressing necessary linguistic context.
    \item \textbf{Adaptive Manifold Re-injection:} By utilizing JSD to regulate the fusion, we effectively re-inject the ``high-precision anchor'' ($P_{do}$) into the ``high-fluency manifold'' ($P_{obs}$). As shown in Figure~\ref{fig:context_loss}, compared to pure attention intervention which suffers from context loss and repetition (red text), our Entropy-Guided Causal Decoding (\M) achieves a superior balance, maintaining factual accuracy (green text) while preserving the natural flow and diversity of the generated response.
\end{itemize}

\subsection{Additional Performance}
\label{app:qualitative_perf}

\begin{figure*}[t]  
    \centering  
    \includegraphics[width=0.9\linewidth]{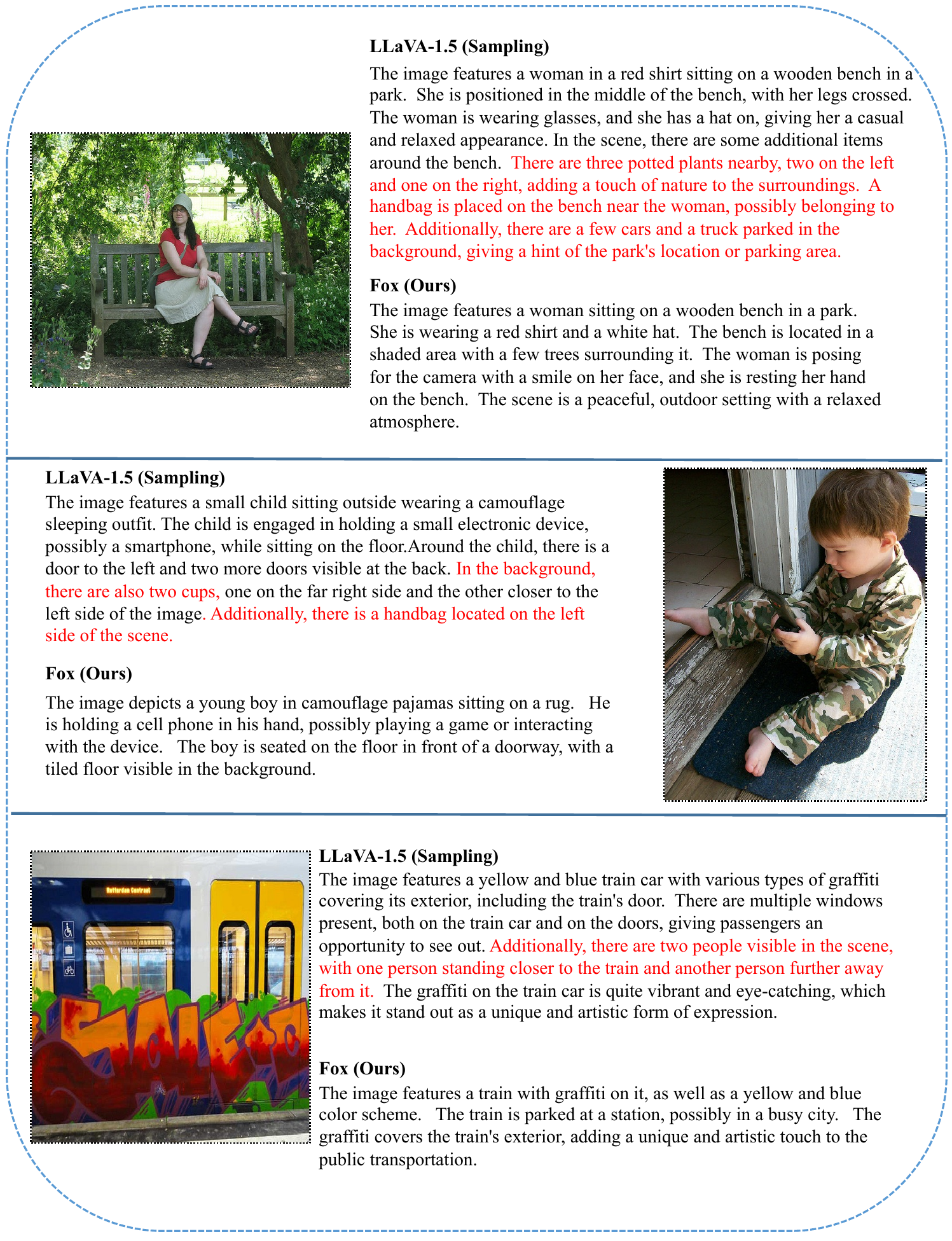}  
    \caption{\M’s performance on reducing hallucinations of LLaVA-1.5.} 
    \label{fig:llava_app}  
\end{figure*}

\begin{figure*}[t]  
    \centering  
    \includegraphics[width=0.9\linewidth]{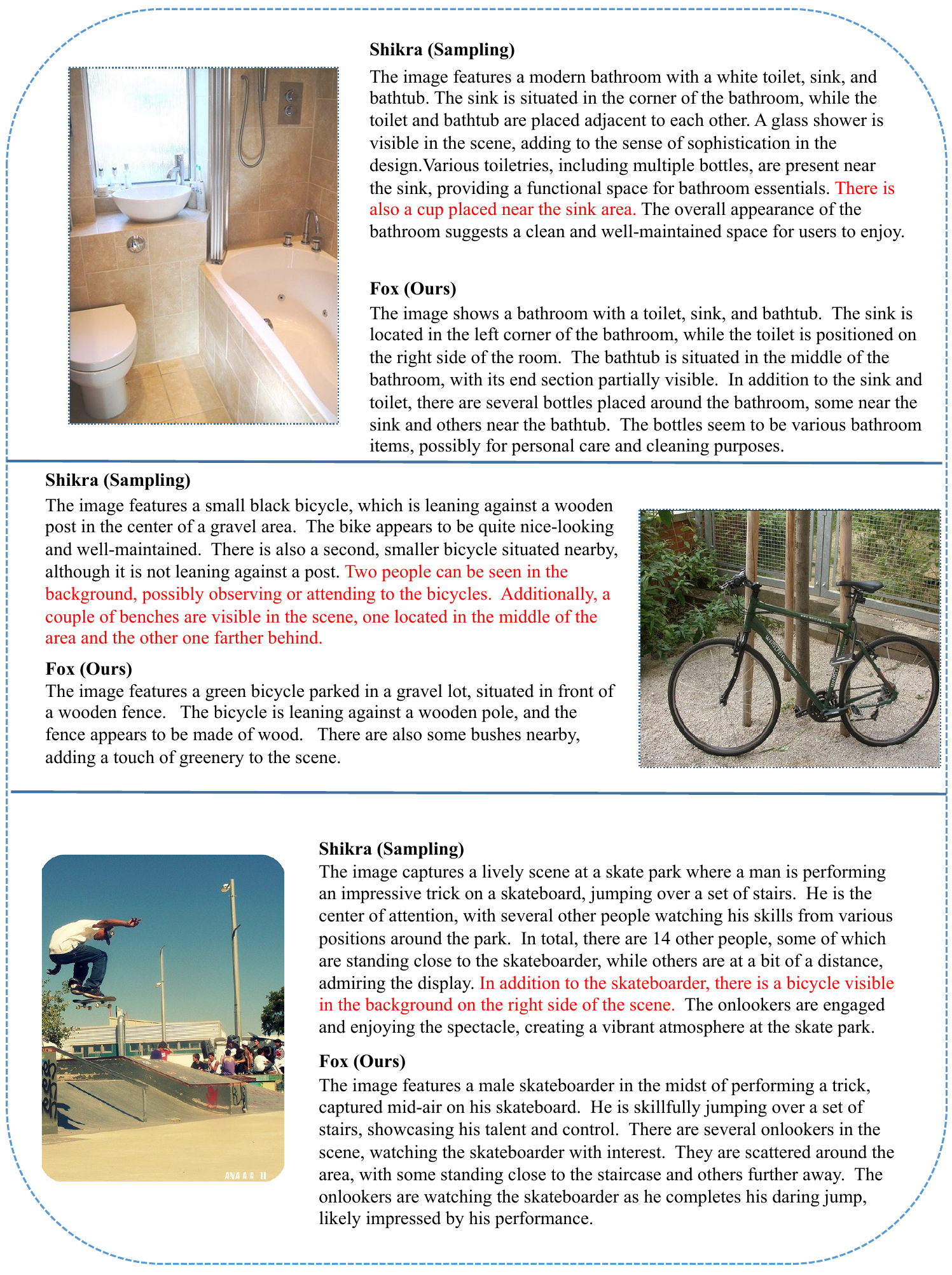}  
    \caption{\M’s performance on reducing hallucinations of InstructBLIP.} 
    \label{fig:shikra_app}  
\end{figure*}

\begin{figure*}[t]  
    \centering  
    \includegraphics[width=0.9\linewidth]{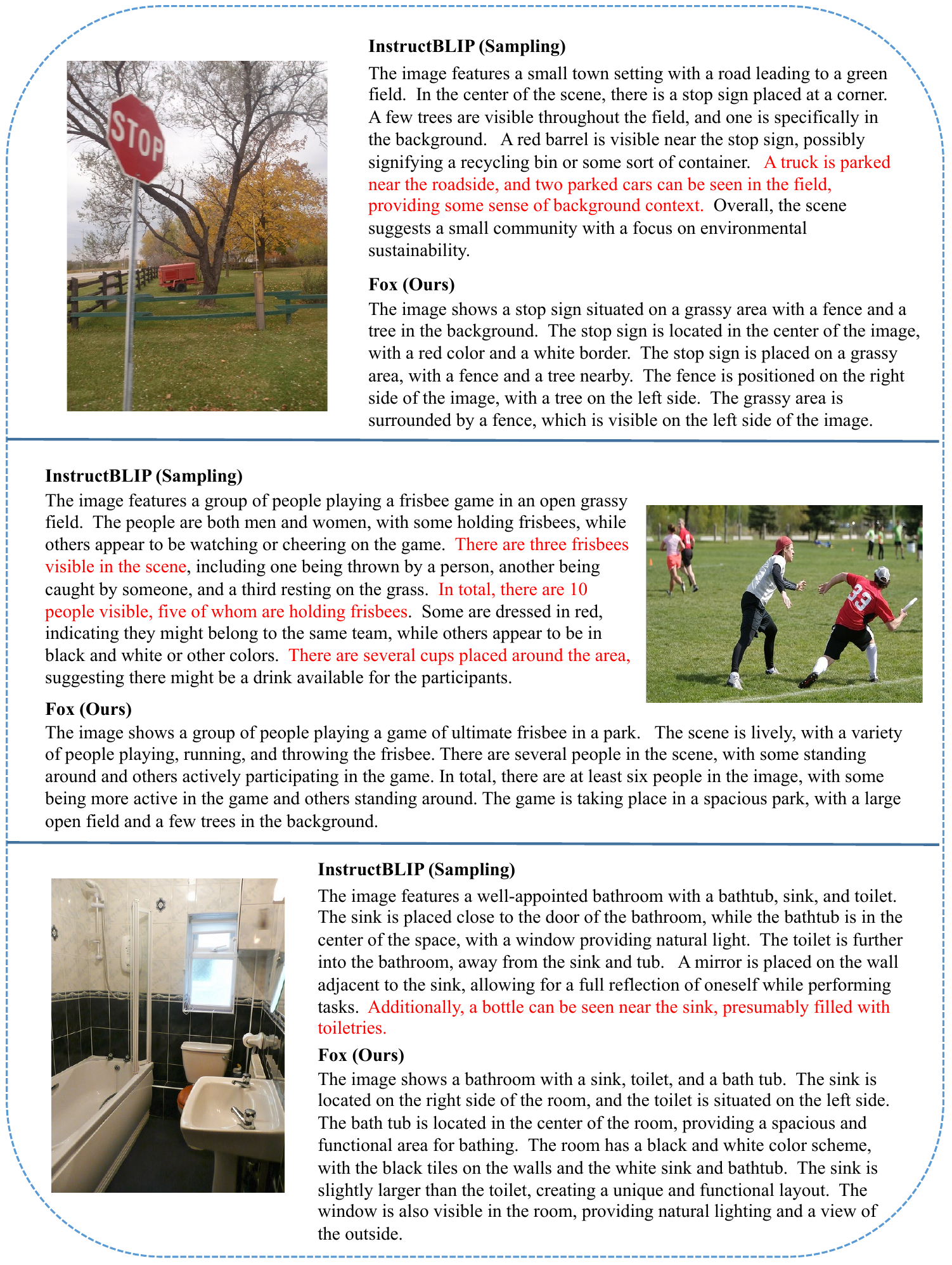}  
    \caption{\M’s performance on reducing hallucinations of Shikra.} 
    \label{fig:instructBLIP-app}  
\end{figure*}

We present additional qualitative examples to showcase the practical performance of \M\ in reducing hallucinations across different LVLM backbones.
As shown in Fig.~\ref{fig:llava_app}, \M\ effectively suppresses prior-driven hallucinated attributes and objects on \textsc{LLaVA-1.5} while preserving visually grounded details.
Fig.~\ref{fig:shikra_app} further demonstrates that the proposed intervention generalizes to \textsc{InstructBLIP}.
Finally, Fig.~\ref{fig:instructBLIP-app} reports additional cases on \textsc{InstructBLIP} and \textsc{Shikra}, where \M\ consistently improves visual grounding under diverse scenes and object configurations.
These examples complement the quantitative results in the main paper by providing intuitive evidence of cross-backbone robustness.

%%%%%%%%%%%%%%%%%%%%%%%%%%%%%%%%%%%%%%%%%%%%%%%%%%%%%%%%%%%%%%%%%%%%%%%%%%%%%%%
%%%%%%%%%%%%%%%%%%%%%%%%%%%%%%%%%%%%%%%%%%%%%%%%%%%%%%%%%%%%%%%%%%%%%%%%%%%%%%%

\end{document}